\documentclass[journal]{IEEEtran}

\ifCLASSINFOpdf
\else
\fi

\usepackage{booktabs}
\usepackage{amsmath}
\usepackage{rotating}
\usepackage[pagebackref=true,breaklinks=true,colorlinks,linkcolor=blue,citecolor=blue]{hyperref}

\usepackage{mathrsfs}
\usepackage[table]{xcolor}

\usepackage{color}
\usepackage{times}
\usepackage{epsfig}
\usepackage{graphicx}
\usepackage{amssymb}
\usepackage{multirow}
\usepackage{algorithm}
\usepackage{algorithmic}
\usepackage{array}
\usepackage{cite}
\newcommand{\RN}[1]{%
  \textup{\uppercase\expandafter{\romannumeral#1}}%
}

\usepackage[utf8]{inputenc}
\usepackage[T1]{fontenc}
\usepackage{textcomp}
\usepackage[table]{xcolor}
\usepackage{threeparttable}

\makeatletter
\def\ps@IEEEtitlepagestyle{%
  \def\@oddfoot{\mycopyrightnotice}%
  \def\@evenfoot{}%
}
\def\mycopyrightnotice{%
  {\hfill  \hfill}
}
\makeatother

\begin{document}

\title{Underwater Object Detection in the Era of Artificial Intelligence: Current, Challenge, and Future}
\author{Long Chen, Yuzhi Huang, Junyu Dong, Qi Xu,  Sam Kwong, Huimin Lu, Huchuan Lu, and Chongyi Li 
\thanks{L. Chen is with the Department of Medical Physics and Biomedical Engineering, University College London, United Kingdom.}
\thanks{Y. Huang and Q. Xu are with the School of Computer Science and Technology, Dalian University of Technology, China.}
\thanks{J. Dong with the Department of Information Science and Engineering, Ocean University of China, China.}
\thanks{S. Kwong is with the School of Data Science, Lingnan University, Hong Kong, China.}
\thanks{H. Lu is with the School of Automation, Southeast University, China, and also with the Advanced Institute of Ocean, Southeast University, China}
\thanks{H. Lu is with School of Information and Communication Engineering, Dalian University of Technology, Dalian, China}
\thanks{C. Li is with VCIP, CS, Nankai University, China. C. Li is the corresponding author.}
}

\markboth{}
{Shell \MakeLowercase{\textit{et al.}}: Bare Demo of IEEEtran.cls for IEEE Journals}

\maketitle
\begin{abstract}

Underwater object detection (UOD), aiming to identify and localise the objects in underwater images or videos, presents significant challenges due to the optical distortion, water turbidity, and changing illumination in underwater scenes. In recent years, artificial intelligence (AI) based methods, especially deep learning methods, have shown promising performance in UOD. To further facilitate future advancements, we comprehensively study AI-based UOD. 
In this survey, we first categorise existing algorithms into traditional machine learning-based methods and deep learning-based methods, and summarise them by considering learning strategy, experimental dataset, utilised features or frameworks, and learning stage. Next, we discuss the potential challenges and suggest possible solutions and new directions. We also perform both quantitative and qualitative evaluations of mainstream algorithms across multiple benchmark datasets by considering the diverse and biased experimental setups. Finally, we introduce two off-the-shelf detection analysis tools, Diagnosis and TIDE, which well-examine the effects of object characteristics and various types of errors on detectors. These tools help identify the strengths and weaknesses of detectors, providing insigts for further improvement. The source codes, trained models, utilised datasets, detection results, and detection analysis tools are public available at \url{https://github.com/LongChenCV/UODReview}, and will be regularly updated.

\end{abstract}


\IEEEpeerreviewmaketitle

\section{Introduction}

\begin{figure}[tb]
\centering
\includegraphics[width=9cm]{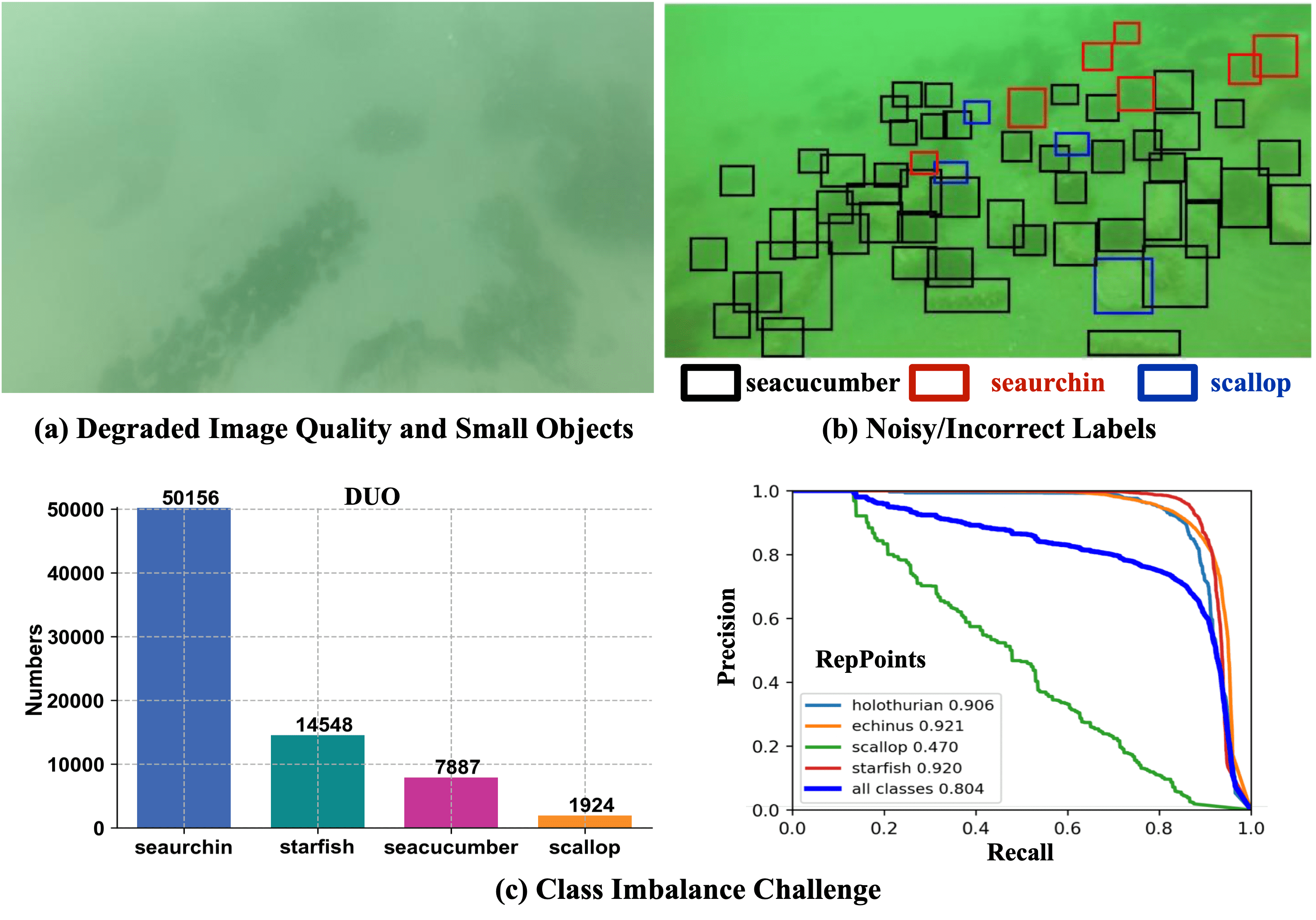}
\caption{Challenges such as (a) degraded image quality and small objects, (b) noisy labels, and (c) class imbalance in underwater object detection datasets significantly impair the performance of deep detection models. For instance, (c) highlights that the deep detector RepPoints \cite{yang2019reppoints} faces severe class imbalance issues on the DUO \cite{liu2021dataset} dataset.}
\label{fig:challenges}
\end{figure}

Underwater object detection (UOD) aims not only to identify object categories but also to predict their locations. It remains one of the most challenging tasks in computer vision due to the complex underwater environment, where the captured images frequently suffer from severe blur, color distortion, and degraded visibility \cite{chiang2011underwater, fu2023rethinking}. As shown in Fig. \ref{fig:challenges}, challenges such as degraded image quality, small object sizes, noisy labels, and class imbalances significantly hinder the performance of underwater object detection models.

\begin{figure}[tb]
\centering
\includegraphics[width=9cm]{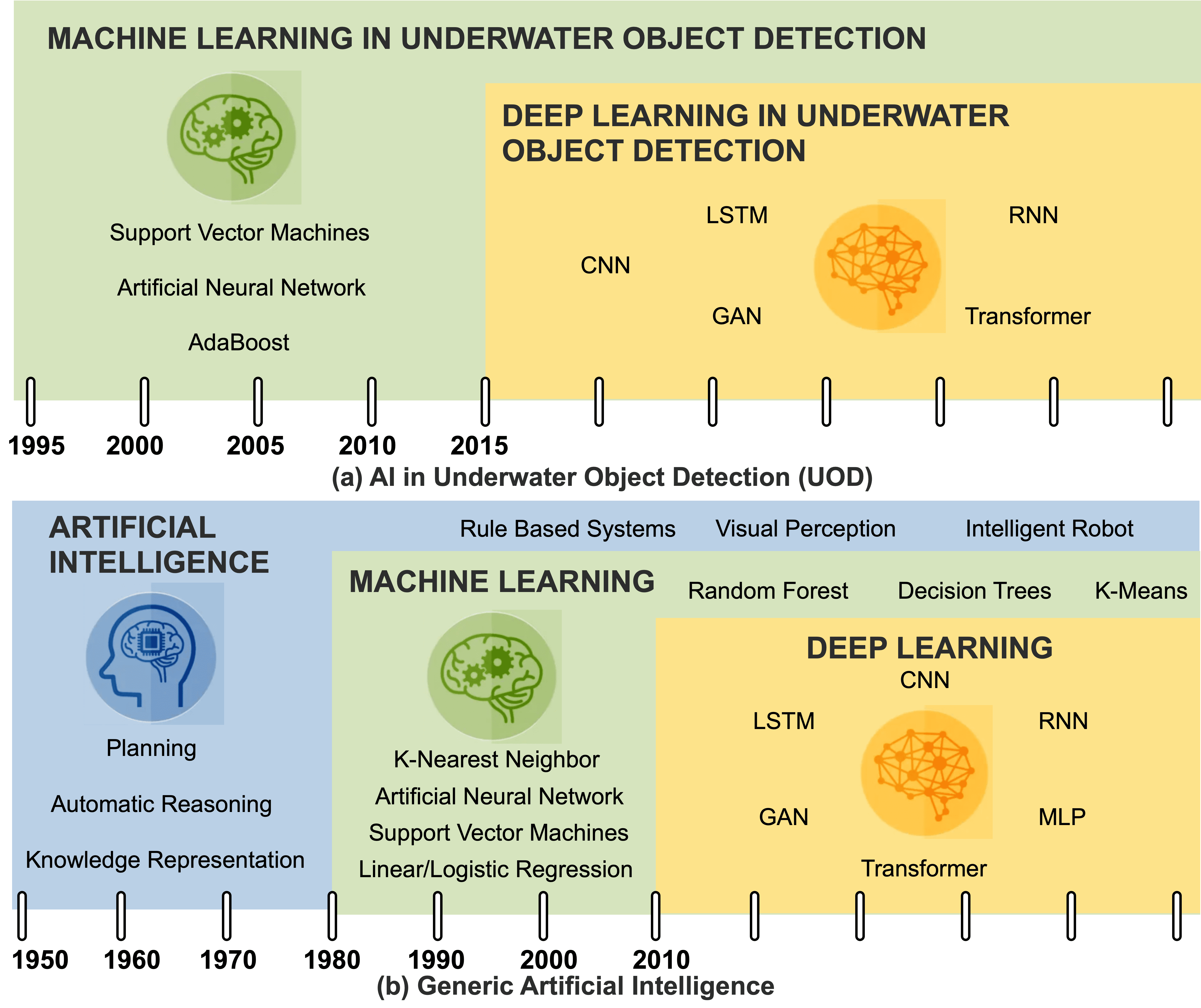}
\caption{The development of (a) AI in underwater object detection and (b) generic artificial intelligence.}
\label{fig:AIDLML}
\end{figure}

Recent years have witnessed an exponential growth in artificial intelligence (AI)-based UOD algorithms, driven by advancements in computational power and data availability \cite{akkaynak2018revised, xu2023systematic}. AI, a general concept of the technique that enables computers to emulate human intelligence, has garnered attention since its inception in the 1950s, as illustrated in Fig \ref{fig:AIDLML}. After an initial wave of optimism, specialised fields of AI-first machine learning, and later deep learning-triggered notable disruptions. By 1995, AI technology began to be integrated into the field of UOD following its revival. Today, AI, particularly its key branch, machine learning, is extensively utilised in the field of UOD. Previous machine learning-based UOD methods can be broadly categorised into traditional machine learning methods and deep learning methods. To gain insight into their strengths and weaknesses, we summarised these methods based on four aspects:
\begin{itemize}
\item \textit{Learning Strategies}
\item \textit{Applicable Datasets}
\item \textit{Utilised Features or Frameworks}
\item \textit{Learning Stages}
\end{itemize}

\begin{figure}[tb]
\centering
\includegraphics[width=6.6cm]{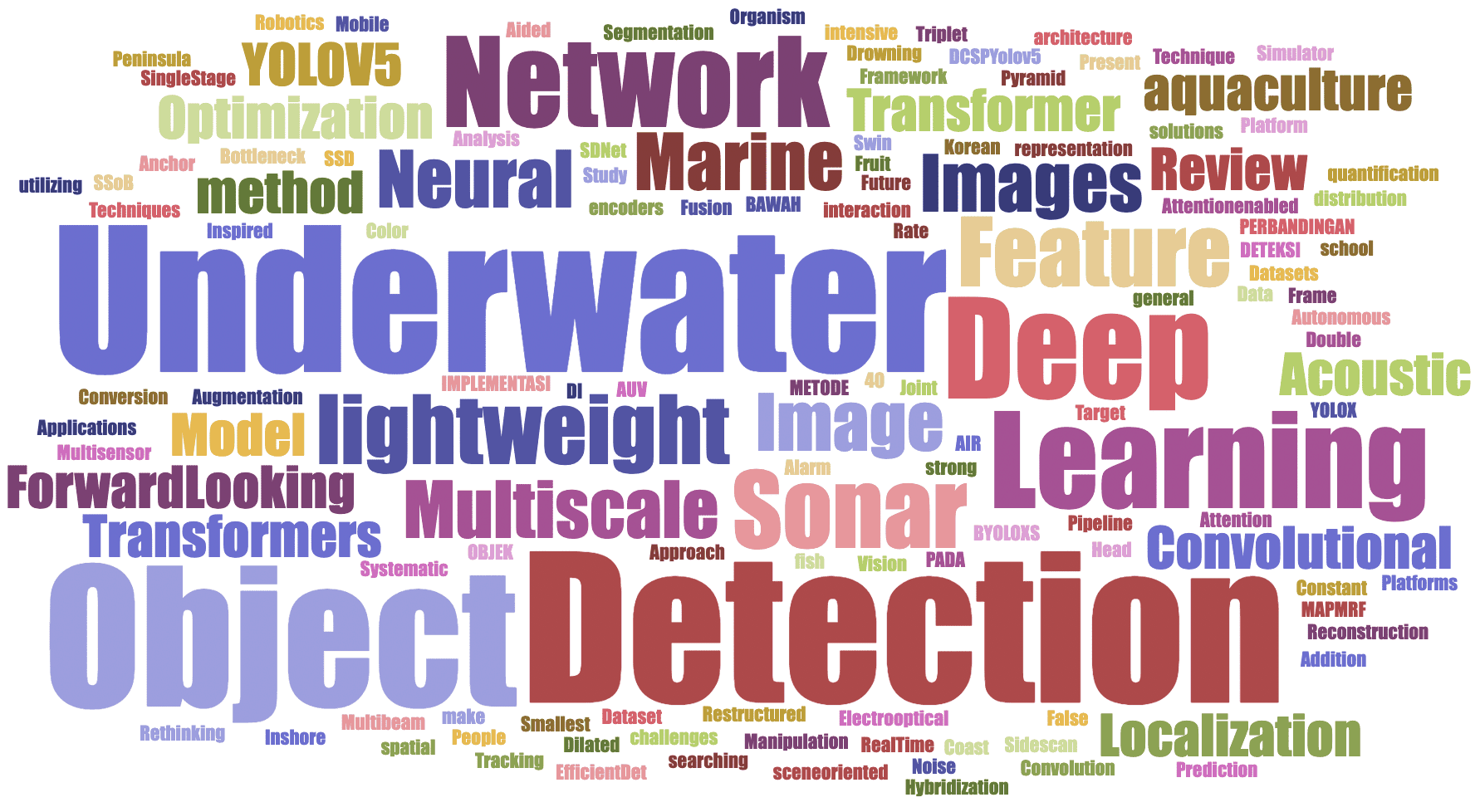}
\caption{The frequency map of the keyword ‘underwater object detection' in Google Scholar from 2020 to 2024. The size of the keyword is proportional to the frequency of the word. The keywords ‘underwater object detection’, ‘deep learning’, and ‘neural network’ have drawn large research interests in the community.}
\label{fig:keyword}
\end{figure}

Among various AI techniques, deep learning has garnered significant attention and extensive study in UOD, as illustrated in Fig. \ref{fig:keyword}. Deep learning models require large amounts of training data but offer clear advantages over traditional methods in terms of detection accuracy. Two main research lines are primarily pursued in UOD. First, general deep detection networks, such as Faster RCNN \cite{ren2016faster}, SSD \cite{liu2016ssd}, YOLO \cite{redmon2016you}, and their variants, have been adapted for underwater object detection, significantly enhancing performance in this field. The second research line focuses on developing special network backbones, loss functions, or learning strategies for UOD. 

Despite the significant potential of deep learning-based UOD techniques for both academic and commercial applications, they still faces serious challenges. Many generic detection networks struggle to achieve accurate detection due to special difficulties unique to underwater domain. To facilitate a comprehensive understanding of these challenges, we categorise them into four groups: 
\begin{itemize}
\item \textit{Image Quality Degradation Challenge}
\item \textit{Small Object Detection Challenge}
\item \textit{Noisy Label Challenge}
\item \textit{Class Imbalance Challenge}
\end{itemize}

During the solution search stage, we have observed significant research gaps between generic object detection (GOD) and UOD in addressing these challenges. To bridge the research gaps and advance UOD development, we summarise current solutions for each challenge in both UOD and GOD and suggest potential improvements and new research directions.

Datasets are the fundamental to learning-based UOD approaches. Over the past decades, numerous datasets have been introduced. These public benchmarks offer platforms for the evaluations of UOD methods and significantly advance related fields. Hence, we review existing UOD datasets and offer comprehensive and detailed descriptions of both the datasets and the evaluation metrics. Moreover, we introduce two valuable detection analysis tools, Diagnosis \cite{hoiem2012diagnosing} and TIDE \cite{bolya2020tide}, to assess the specific strengths and weaknesses of detectors by analysing their errors. Diagnosis \cite{hoiem2012diagnosing} 
 well-examines how object characteristics, such as size and aspect ratio, impact the detection performance, while TIDE  \cite{bolya2020tide} well-evaluates the effects of different types of errors on detectors. 

\begin{table*}[ht]
\label{tab:com_surveys}
\begin{center}
\caption{The summary of the existing underwater object detection surveys. The quality of each survey is rated based on five key criteria: review of traditional learning-based UOD, review of deep learning-based UOD, datasets, challenges, and proposed solutions. More stars ($\bigstar$) indicate better quality.}
\label{tab:uodreview1}
\begin{tabular}{p{0.5cm}p{4.5cm}p{3.5cm}p{1.2cm}p{1.2cm}p{1.1cm}p{1cm}p{1cm}p{1cm}}
\toprule
Year & Title & Venue & ML UOD & DL UOD & Dataset & Challenge & Future\\
\hline
2017 & Deep Learning on Underwater Marine Object Detection: A Survey \cite{moniruzzaman2017deep} & (Conference) Advanced Concepts for Intelligent Vision Systems 2017 & $\bigstar$ & $\bigstar\bigstar$ & $\bigstar$ & $\bigstar$ & $\bigstar$ \\
2020 & Robust Underwater Object Detection with Autonomous Underwater Vehicle: A Comprehensive Study \cite{gomes2020robust} & (Conference) Proceedings of the International Conference on Computing Advancements & $\bigstar\bigstar\bigstar$ & $\bigstar\bigstar\bigstar$ & $\bigstar$ & $\bigstar$ & $\bigstar$\\
2022 & Review on deep learning techniques for marine object recognition: Architectures and algorithms \cite{wang2022review} & (Journal) Control Engineering Practice & $\bigstar$ & $\bigstar$ & $\bigstar\bigstar$ & $\bigstar$ & $\bigstar$\\
2022 & Underwater Object Detection: Architectures and Algorithms–A Comprehensive Review \cite{fayaz2022underwater} & (Journal) Multimedia Tools and Applications & $\bigstar$ & $\bigstar$ & $\bigstar$ & $\bigstar$ & $\bigstar$\\
2023 & A Systematic Review and Analysis of Deep Learning-based Underwater Object Detection \cite{xu2023systematic} & (Journal) Neurocomputing & $\bigstar$ & $\bigstar\bigstar\bigstar$ & $\bigstar$ & $\bigstar\bigstar\bigstar$ & $\bigstar\bigstar$\\
2023 & Rethinking General Underwater Object Detection: Datasets, Challenges, and Solutions \cite{fu2023rethinking} & (Journal) Neurocomputing & $\bigstar$ & $\bigstar\bigstar$ & $\bigstar\bigstar\bigstar$ & $\bigstar\bigstar\bigstar$& $\bigstar\bigstar$ \\
2024 & Underwater Object Detection and Datasets: A Survey \cite{jian2024underwater} & (Journal) Intelligent Marine Technology and Systems & $\bigstar\bigstar$ & $\bigstar\bigstar$ & $\bigstar\bigstar$ & $\bigstar\bigstar$ & $\bigstar$\\
2024 & Ours & - & $\bigstar\bigstar\bigstar$ & $\bigstar\bigstar\bigstar$ & $\bigstar\bigstar\bigstar$ & $\bigstar\bigstar\bigstar$ & $\bigstar\bigstar\bigstar$\\
\hline
\bottomrule
\end{tabular}
\end{center}
\end{table*}

\begin{figure*}[htb]
\centering
\includegraphics[width=17cm]{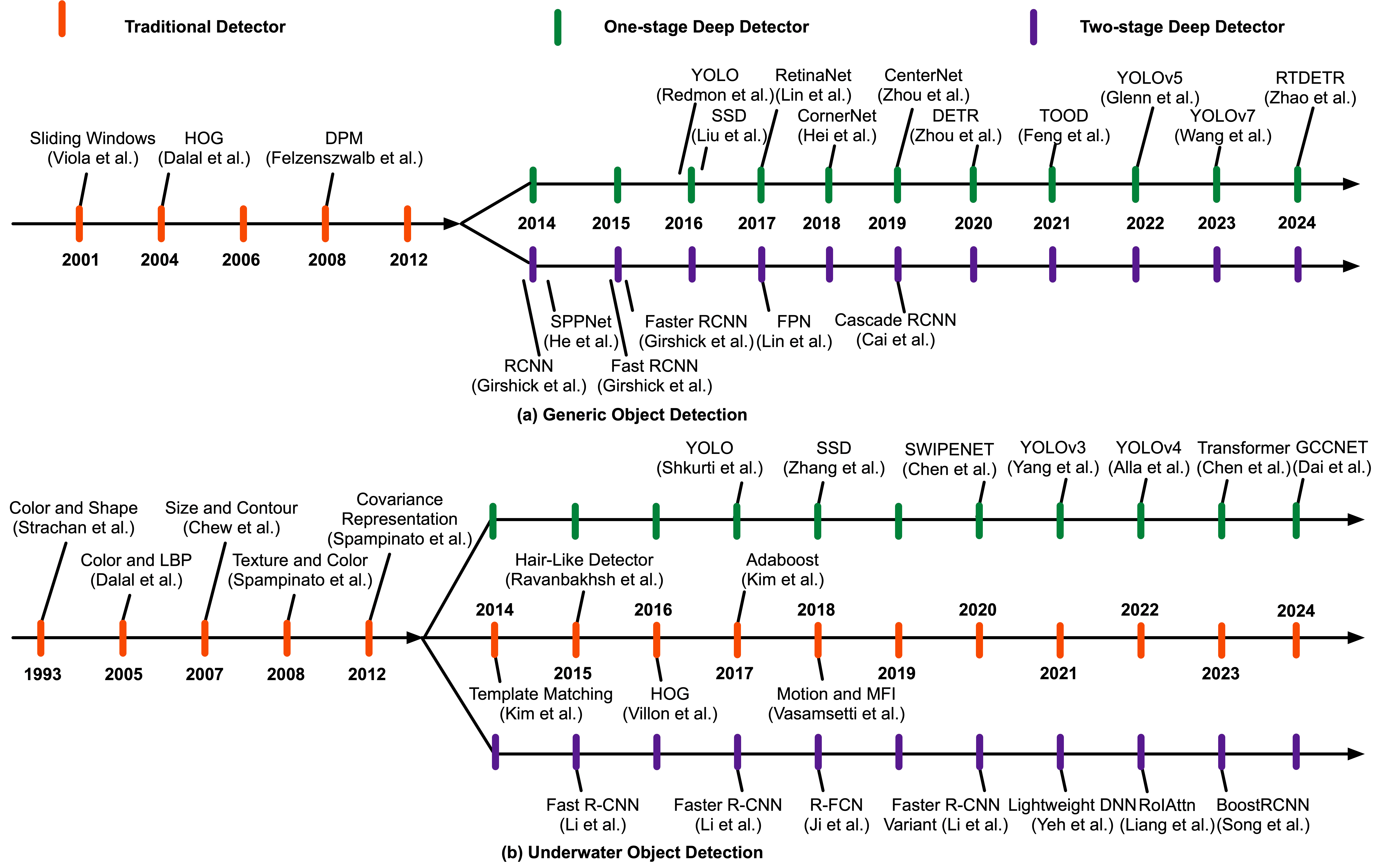}
\caption{The road maps of (a) generic object detection and (b) underwater object detection reveal that the latter often leverages insights and techniques from the former to enhance detection performance in underwater envrionments.}
\label{fig:roadmap}
\end{figure*}
\subsection{Comparisons with Previous Surveys}
\label{sec:previous_surveys}
In the last few years, quite a number of surveys on underwater object detection have been published. For this work, we cautiously select seven representative survey papers \cite{moniruzzaman2017deep, gomes2020robust, wang2022review, fayaz2022underwater, xu2023systematic, fu2023rethinking, jian2024underwater} for comparison, most of which are the latest surveys published in last three years. To quantitatively assess the quality of the selected surveys, we evaluate them across five key aspects: coverage of traditional machine learning-based UOD, deep learning-based UOD, datasets, challenges, and proposed solutions. We posit that a comprehensive survey of underwater object detection should provide researchers with a clear understanding of the field's potential, challenges and future directions. To quickly convey the strengths and weaknesses of each survey paper, we rate the quality of each aspect using a star system $\bigstar$, with more stars indicating higher review quality. The quantitative evaluations are presented in Table \ref{tab:com_surveys}, and detailed discussions of each paper follow.

In 2017, Moniruzzaman et al. \cite{moniruzzaman2017deep} classified the underwater object detection approaches according to the object categories. While the paper offers a comprehensive reviews on deep learning applications in fish, plankton, and coral recognition, it lacks analysis of the challenges and potential future solutions for underwater object detection. Gomes et al. \cite{gomes2020robust} conducted a broad study on traditional machine learning and deep learning approaches for underwater object detection. Although they review and compare a wide range of methods, the analysis of challenges and proposed solutions remains limited. Wang et al. \cite{wang2022review} provided comprehensive discussions on deep learning techniques for general object recognition and detection, but dedicated minimal attention to underwater object detection. Similarly, Fayaz et al. \cite{fayaz2022underwater} concentrated heavily on the development of general deep learning methods, overlooking the specific techniques used for underwater object detection. Xu et al. \cite{xu2023systematic} provided a comprehensive review of the potential applications and research challenges in underwater object detection. This work also examines the connection between underwater image enhancement and underwater object detection, providing some insights for future research directions in the field. Fu et al. \cite{fu2023rethinking} presented a review focused on deep learning-based underwater object detection, but traditional machine learning methods haven't been covered. This paper also provides a comprehensive survey of datasets and challenges, and suggests some promising solutions to advance underwater object detection. Jian et al. \cite{jian2024underwater} presented a moderate review of traditional machine learning and deep learning-based methods, along with discussions on datasets and challenges. However, their review offers limited insights for further advancements.

Table \ref{tab:com_surveys} reveals several common issues in previous surveys. Firstly, many survey papers, such as \cite{akkaynak2018revised, xu2023systematic, fu2023rethinking}, exclusively focus on an intensive review on deep learning-based UOD techniques, however, neglecting the reviews on traditional machine learning approaches. Several works \cite{wang2022review, fayaz2022underwater} overemphasise generic object detection methods but ignore the core topic of underwater object detection. Secondly, most studies, such as \cite{moniruzzaman2017deep, gomes2020robust, wang2022review, fayaz2022underwater, jian2024underwater}, offer limited explorations of challenges and corresponding solutions for UOD. Therefore, this survey focuses on discussing these challenges and their potential solutions in depth.

\subsection{Contributions of This Survey}

In this paper, we provide a more comprehensive review of AI-based UOD techniques and discuss key challenges, potential solutions, available datasets, evaluation metrics, and useful analysis tools for underwater object detection. The contributions of this survey are summarised as follow:

\begin{itemize}
\item This survey focuses on reviewing AI techniques in underwater object detection, unlike previous work that centers on generic object detection. These UOD techniques are categorised based on learning strategies, applicable datasets, utilised features or network architectures, and learning stages, providing a clearer understanding of their strengths and weaknesses.

\item  An in-depth analysis of the challenges in underwater object detection is presented, with these challenges summarised into four categories: image quality degradation, noisy labels, small object detection, and class imbalance. Advanced techniques from both generic and underwater object detection are explored and summarised to provide potential solutions to these challenges, offering valuable insights for future development. 

\item A comprehensive overview of underwater object detection datasets and evaluation metrics is provided. Moreover, two valuable detection analysis tools, Diagnosis and TIDE, are introduced to identify the strengths and weaknesses of each detector by analysing their errors. The pre-built source code for these tools, tailored for the RUOD and DUO datasets, are available online.

\item To ensure a fair evaluation of underwater detectors, we recommend two high-quality and large-scale benchmarks, RUOD and DUO, and evaluate several mainstream deep detectors on these datasets. The source code, trained models, utilised datasets, and detection results are available online, offering researchers an accessible platform to compare their detectors against previous works. 

\end{itemize}

The remainder of this paper is structured as follows: Section \ref{sec:related_works} reviews related works on AI-based UOD methods. Section \ref{sec:challenge_solution} describes the research challenges in UOD and suggests potential solutions. Section \ref{sec:datasets} presents popular datasets, evaluation metrics, and introduces two valuable detection analysis tools. Section \ref{sec:experiments} evaluates several mainstream detection frameworks on two large-scale datasets, RUOD and DUO, and reports and discusses the experimental results. Finally, future insights and vision are provided in Section \ref{sec:conclusions}.

\section{Existing Works of AI-based UOD}
\label{sec:related_works}

The development of underwater object detection is closely tied to the advancements in generic object detection, as evidenced by the road maps of GOD and UOD presented in Fig. \ref{fig:roadmap}. Many generic object detection frameworks and algorithms have been directly applied to underwater object detection and significantly advanced the research field. Hence, we first review the progress in GOD to identify similarities and gaps between the two areas. Then, we review related works of AI-based UOD methods, including both traditional machine learning methods and deep learning methods. Table \ref{tab:reviewmachine} and Table \ref{tab:reviewdeep} summarise the traditional and deep learning-based UOD approaches, detailing their learning strategies, utilised datasets, features/techniques or detection frameworks used.

\subsection{Generic Object Detection}

The evolution of generic object detection can be divided into two periods: the traditional era (before 2014) and the deep learning era (after 2014). 

\subsubsection{Traditional Generic Object Detection}

During the traditional period, researchers focused on developing complex hand-crafted features for object detection. In 2001, Viola and Jones \cite{viola2001rapid} introduced a real-time face detection algorithm that used sliding windows to scan all possible locations in an image and applied a detector to identify whether the window contained a human face. In 2005, Dalal et al. \cite{dalal2005histograms} introduced an effective hand-crafted feature called Histogram of Rriented Gradients (HOG) for pedestrian detection. In 2008, Felzenszwalb et al. \cite{felzenszwalb2008discriminatively} developed the Deformable Part-Based Model (DPM) for generic object detection, which became the champion model in the Pascal VOC detection challenge. However, by 2010, traditional object detection methods reached a plateau as the performance of hand-crafted features became saturated.

\subsubsection{Deep Generic Object Detection}

It is widely recognised that the deep learning era in object detection began with the introduction of the two-stage deep detection network, RCNNs \cite{girshick2014rich}, proposed by Girshick et al. in 2014. RCNNs have since served as a crucial foundation for numerous deep detection networks detectors. Following 2014, the filed of generic object detection evolved at an unprecedented pace. There are two main research lines in deep learning-based object detection: two-stage deep detectors and one-stage deep detectors.

\textit{Two-stage Detectors.} Two-stage detectors, such as Faster RCNN \cite{ren2016faster}, RFCN \cite{dai2016r}, and FPNs \cite{lin2017feature}, initially employ a proposal generation technique to produce a set of object proposals. Subsequently, a separate deep network extracts features from these proposals and predicts their locations and categories. The milestone work, Faster RCNN, integrates proposal generation, feature extraction, and bounding box regression into a unified end-to-end framework. Since then, the end-to-end deep pipeline has become dominant in object detection research. These two-stage deep detectors have significantly advanced detection performance in authoritative object detection challenges, such as the ImageNet and Microsoft COCO. However, many of these detectors struggle to achieve real-time detection due to their complex, coarse-to-fine processing paradigm.

\textit{One-stage Detectors.} To overcome speed limitations, one-stage detectors like YOLO \cite{redmon2016you} and SSD \cite{liu2016ssd} were introduced to perform object detection in real-time. Although these detectors excel in speed, they initially lagged behind two-stage detectors in accuracy. However, advanced one-stage frameworks such as RetinaNet \cite{lin2017focal}, CornerNet \cite{law2018cornernet}, and CenterNet \cite{duan2019centernet} later emerged, surpassing two-stage detectors in both accuracy and speed. In recent years, transformer-based networks such as DETR \cite{carion2020end} and Deformable DETR \cite{zhu2020deformable} have attracted considerable attention in GOD. These models utilise attention mechanisms instead of convolution layers, which provide a global receptive field and show great promise for future advancement.

From the road maps presented in Fig. \ref{fig:roadmap}, it is clear that underwater object detection follows a similar development trajectory to generic object detection. Many techniques from generic object detection have been adapted to enhance underwater detection performance. Therefore, we believe that advanced techniques from generic object detection can benefit underwater object detection, and vice versa, in the further. 

\subsection{Traditional ML-based Underwater Object Detection}

Traditional machine learning algorithms have been employed in underwater object detection task for a long history. These approaches generally consists of two-stage learning. In the first stage, hand-crafted features are extracted, which can include simple visual features (e.g., color \cite{kim2014artificial} and shape \cite{strachan1993recognition}) or complex hand-crafted features (e.g., HOG, SIFT and SURF). In the second stage, these features are forwarded to traditional classifiers, such as SVM and decision trees, to carry out various tasks in underwater scenes. Since traditional machine learning-based methods encompass a range of diverse techniques, we categorise them into sonar-based and RGB-based UOD methods based on the data used, and summarise their advantages and disadvantages.
\begin{figure}[htb]
\centering
\includegraphics[width=9cm]{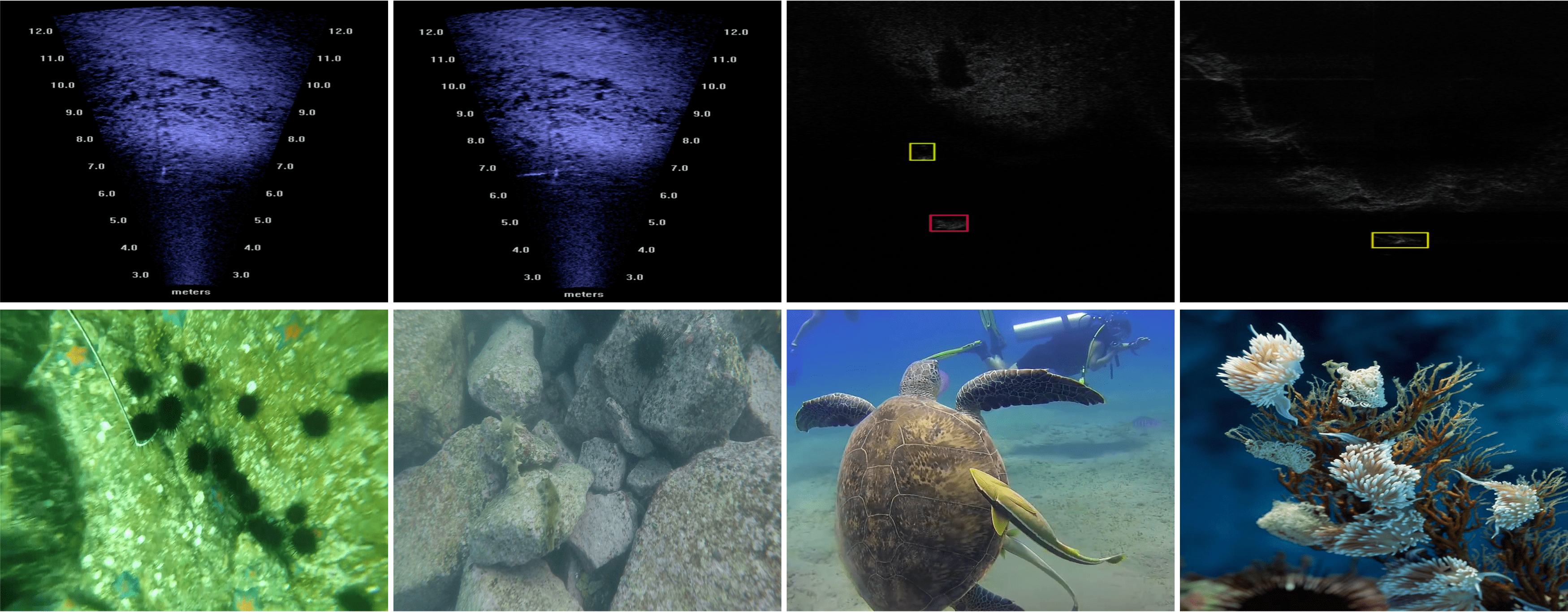}
\caption{Comparisons of the sonar (top) and RGB (bottom) images. RGB images capture rich visual features but with limited perceptual range. Sonar images extend the perceptual range but are less intuitive and harder for humans to interpret.}
\label{fig:camera_sonar}
\end{figure}

\begin{table*}
\begin{center}
\caption{Summary of Traditional Machine Learning Approaches for UOD, focusing on learning strategies (LS), utilised datasets, input data formats, extracted features, and detection methods.}
\label{tab:reviewmachine}
\begin{tabular}{p{1.2cm}p{0.7cm}p{0.7cm}p{3cm}p{0.7cm}p{4cm}p{4cm}}
\toprule
Refs & Year & LS & Datasets & Format & Features & Detecotrs \\
\hline
Strachan et al. \cite{strachan1993recognition} & 1993 & SL & Private Dataset & RGB & Color and Shape Features & Discriminant Analysis\\
\hline
Shiela et al. \cite{marcos2005classification} & 2005 & SL & Great Barrier Reef Dataset & RGB & Color and Local Binary Pattern (LBP) Features & Artificial Neural Network (ANN)\\
\hline
Barat et al. \cite{barat2006robust} & 2006 & UL & Private Dataset & RGB & Visual Attention Mechanisms & Motion-based Algorithm\\
\hline
Chew et al. \cite{chew2007automatic} & 2007 & UL & Private Dataset &  Sonar &  Frequency and Contour Features & Self-adaptive Power Filtering Technique\\
\hline
Spampinato et al. \cite{spampinato2008detecting}  & 2008 & UL & Private Dataset & RGB & Texture and Color Features & Moving Average Algorithm\\
\hline
Beijbom et al. \cite{beijbom2012automated} & 2012 & SL & Moorea Labeled Corals Dataset & RGB & Texture and Color Features & Support Vector Machine (SVM)\\
\hline
Spampinato et. al \cite{spampinato2012covariance} & 2012 & UL & Private Dataset & RGB & Covariance Representation & Covariance Tracking Algorithm\\
\hline
Galceran et al. \cite{galceran2012real} & 2012 & UL & Private Dataset & Sonar & Integral-Image Representation & Echo Scoring and Thresholding\\
\hline
Li et al. \cite{li2013underwater} & 2013 & UL & Private Dataset & Sonar & Color and Area Features & Otsu Segmentation Algorithm and Contour Detection Algorithm\\
\hline
Lee et al. \cite{lee2004contour} & 2014 & UL & Private Dataset & RGB & Contour and Shape Features & Contour Matching Algorithm\\
\hline
Kim et al. \cite{kim2014artificial} & 2014 & UL & Private Dataset & RGB & Artificial Landmark Features & Template Matching Algorithm\\
\hline
Ravanbakhsh et al. \cite{ravanbakhsh2015automated} & 2015 & SL & Private Dataset & RGB & Shape Features Modeled by PCA & Haar-Like Detector\\
\hline
Cho et al. \cite{cho2015acoustic} & 2015 & UL & Private Dataset & Sonar & Motion Features of Acoustic Beam Profiles & Analysis the Cross-correlations between Successive Sonar Images\\
\hline
Hou et al. \cite{hou2016underwater} & 2016 & UL & Private Dataset & RGB & Color and Shape Features & Shape Signature Algorithm\\
\hline
Chuang et al. \cite{chuang2016feature} & 2016 & SL & Fish4Knowledge and NOAA Fisheries Datasets & RGB & Part Features & An Error-Resilient Classifier\\
\hline
Villon et al. \cite{villon2016coral} & 2016 & SL & MARBEC Dataset & RGB & HOG Features & Support Vector Machine (SVM)\\
\hline
Liu et al. \cite{liu2016combining} & 2016 & UL & Private Dataset & RGB & Motion Features & Background Subtraction and Three Frame Difference algorithms\\ 
\hline
Chen et. al \cite{chen2017monocular} & 2017 & UL & Private Dataset & RGB & Color, Intensity and Light Transmission Features & Bottom-up ROI Detection and Otsu Segmentation Algorithm\\
\hline
Kim et al. \cite{kim2017imaging} & 2017 & SL & Private Dataset & Sonar & Haar-Like Feature & AdaBoost\\
\hline
Vasamsetti et al. \cite{vasamsetti2018automatic} & 2018 & UL & Fish4Knowledge Dataset & RGB & Color, Motion and Multi Frame Triplet Pattern (MFTP) Features & Multi-Feature Integration (MFI) Framework\\
\hline
\bottomrule
\end{tabular}
\end{center}
\end{table*}

\subsubsection{Sonar Data-based ML UOD Methods} 

Underwater scenes often present significant decrease in visibility because the signals received by sensors are absorbed and distorted by water bodies. As a result, sonar sensors \cite{li2013underwater, cho2015acoustic} have been widely applied in underwater exploration, as they can provide relatively reliable scene data regardless of visibility. Sonar sensors are adept at capturing geometric structure information and can offer insights into underwater scenes even in low-visibility conditions. 

Two main types of sonars are commonly used in sonar-based UOD: side-scan sonar (SSS) \cite{chew2007automatic, yu2021real, barngrover2015brain} and multi-beam forward-looking sonar (FLS) \cite{galceran2012real,zhang2022target, zhou2022automatic}. SSS provides long-range, high-resolution data, allowing for detection across vast survey areas (hundreds of meters long). In contrast, FLS is sutied for closer, more detailed inspection of specific underwater object locations. Chew et al. \cite{chew2007automatic} utilised simple visual features to detect man-made objects in side-scan sonar images, whereas Yu et al. \cite{yu2021real} developed a TR-YOLOv5s network for detecting shipwrecks and containers in side-scan sonar images. Hayes et al. \cite{hayes1992broad} demonstrated that the high-resolution imagery from SSS is effective for identifying potential objects on the seafloor over vast surveyed areas. 

However, for a more detailed inspection of detected targets, FLS is a more appropriate option. Galceran et al. \cite{galceran2012real} introduced a real-time underwater object detection algorithm designed to detect man-made objects in images captured by multi-beam forward-looking sonars. This work used integral-image representation to extract features without the need for training data, significantly reducing computational overload by processing smaller portions of the underwater images. Zhang et al. \cite{zhang2022target} also developed an enhanced YOLOv5 network, called CoordConv-YOLOv5, for underwater object detection in forward-looking sonar images.

Sonar images extend the perceptual range but are less intuitive and harder for human to interpret due to the absence of visual features, as illustrated in Fig. \ref{fig:camera_sonar}. Additionally, sonar images often contain a significant amount of noise, which makes it challenging to ensure the reliability of sonar image recognition and analysis.

\subsubsection{RGB Data-based ML UOD Methods.} 

In contrast to sonars, cameras can capture a large number of RGB images with high spatial and temporal resolutions. We broadly categorise RGB data-based ML UOD approaches into three groups: methods based on static hand-crafted features, methods based on dynamic motion features, and methods-based on innovative pipelines.

\textit{Static Hand-crafted Features-based Methods.} In RGB data-based ML UOD approaches, the most significant research focus has been on developing robust hand-crafted features. In 1993, Strachan et al. \cite{strachan1993recognition} used color and shape features to identify fish transported on a conveyor belt monitored by a digital camera. Later, Beijbom et al. \cite{beijbom2012automated} utilised texture and color features and employed Support Vector Machine (SVM) classifier to detect corals at various scales. In 2015, Ravanbakhsh et al. \cite{ravanbakhsh2015automated} introduced Principal Component Analysis (PCA) to model fish shape features, which were then processed by a Haar-like detector to identify fish heads and snouts. Several traditional machine learning algorithms have employed more sophisticated hand-crafted features such as SIFT \cite{blanc2014fish}, SURF \cite{bay2006surf}, Haar-like feature \cite{kim2017imaging}, HOG \cite{villon2016coral}, and light transformation features \cite{chen2017monocular} for UOD. 

\textit{Dynamic Motion Features-based Methods.} In addition to detecting the objects in still images, some research efforts \cite{barat2006robust, spampinato2012covariance} focus on using motion information to detect moving objects. Spampinato et al. \cite{spampinato2012covariance} introduced a covariance-based tracking algorithm for fish detection in 2012. In 2016, Liu et al. \cite{liu2016combining} exploited background subtraction to handle lighting changes and a three frame difference algorithm to address background noise in moving object detection. Subsequently, Vasamsetti et al. \cite{vasamsetti2018automatic} developed a novel spatiotemporal texture feature for detecting moving objects. Their method demonstrated significant improvements in performance and achieves SOAT results on the Fish4Knowledge dataset.

\textit{Innovative Pipeline-based Methods.} In addition to developing static hand-crafted features and dynamic motion features, many researchers have explored new pipelines or techniques to construct underwater object detection systems. In 2008, Spampinato et al. \cite{spampinato2008detecting} developed a vision pipeline for detecting, tracking, and counting fish in low-quality underwater videos. Lee et al. \cite{lee2004contour} utilised contour matching to recognise fish in fish tanks, while Kim et al. \cite{kim2014artificial} applied multi-template object selection and color-based image segmentation for underwater object detection. For detecting man-made underwater objects, Hou et al. \cite{hou2016underwater} developed a color-based extraction algorithm to identify objects of interest after addressing non-uniform illumination, then an improved Otsu algorithm is employed to eliminate color noise in the backgrounds. The pipeline concluded with a shape signature algorithm to recognise objects based on their shape.

\begin{table*}\scriptsize
\begin{center}
\caption{Summary of Deep Learning Approaches for UOD, focusing on learning strategies (LS), utilised datasets, input data formats, learning stages, novel techniques, and detection frameworks.}
\label{tab:reviewdeep}
\begin{tabular}{p{1.2cm}p{0.7cm}p{0.7cm}p{3cm}p{0.7cm}p{0.7cm}p{4.3cm}p{3cm}}
\toprule
Refs & Year & LS & Datasets & Formats & Stages & Novel Techniques & Detection Frameworks \\
\hline
Li et al. \cite{li2015fast} & 2015 & SL & Datasets from LifeCLIEF Fish Task of ImageCLIEF & RGB & Two & - & Fast RCNN\\
\hline
Villon et al. \cite{villon2016coral} & 2016 & SL & MARBEC Dataset & RGB & Two & Motion from Previous Sliding Window & RCNN invariant\\
\hline
Shkurti et al. \cite{shkurti2017underwater}  & 2017 & SL & Aqua and Synthetic Datasets & RGB & One & - & YOLO Variant\\
\hline
Li et al. \cite{li2017deep} & 2017 & SL & Datasets from LifeCLIEF Fish Task of ImageCLIEF & RGB & Two & - & Faster RCNN Variant\\
\hline{}
Ji et al. \cite{ji2018design} & 2018 & SL & URPC2017 Dataset & RGB & Two & - & R-FCN Variant\\
\hline
Zhang et al. \cite{zhang2018single} & 2018 & SL & URPC2017 Dataset & RGB & One & Multi-scale Features and Context Information & SSD Variant\\
\hline
Lee et al. \cite{lee2018deep} & 2018 & SL & Private and Synthetic Datasets & Sonar & Two & Style Transfer Algorithm for Sonar Image Synthesis & Faster RCNN\\
\hline
Pedersenet al. \cite{pedersen2019detection} & 2019 & SL & Brackish Dataset & RGB & One & - & YOLOv2 and YOLOv3\\
\hline
Chen et al. \cite{long2020underwater} & 2020 & SL & URPC2017 and URPC2018 Datasets & RGB & One & Invert Multi-class Adaboost Algorithm to Handle Noisy Data & SWIPENET (SSD Variant)\\
\hline
Zhang et al. \cite{zhang2020research} & 2020 & SL & URPC2019 Dataset & RGB & One & Image Enhancement Technique & SSD\\
\hline
Chen et al. \cite{chen2020underwater} & 2020 & SL & URPC2017 Dataset & RGB & Two & Mixed Attention Mechanism and Multi-Enhancement Strategy & Faster RCNN Variant\\
\hline
Wang et al. \cite{wang2020yolo} & 2020 & SL & URPC2019 Dataset & RGB & One & - & YOLO Nano\\
\hline
Liu et al. \cite{liu2020towards} & 2020 & SL & URPC2019 Dataset & RGB & One & Data Augmentation Method Water Quality Transfer (WQT) & DG-YOLO\\
\hline
Fan \cite{fan2020dual} & 2020 & SL & UWD Dataset & RGB & One & Multi-scale Contextual Features and Anchor Refinement & FERNet (SSD Variant)\\
\hline
Zhang et al. \cite{zhang2020mffssd} & 2020 & SL & URPC Dataset & RGB & One & Attention Module and Multi-scale Feature Fusion & MFFSSD\\
\hline
Lin \cite{lin2020roimix} & 2020 & SL & URPC2018 Dataset & RGB & Two & Data Augmentation Technique RoIMix & Faster RCNN Variant\\
\hline
Karimanzira et al. \cite{karimanzira2020object} & 2020 & SL & Private Dataset & Sonar & Two & - & Faster RCNN Variant\\
\hline
Sung et al. \cite{sung2020realistic} & 2020 & SL & Private and Synthetic Datasets & Sonar & One & Synthesising Sonar Images with GANs & YOLO\\
\hline
Yang et al. \cite{yang2021research} & 2021 & SL & URPC2017 Dataset & RGB & One & - & YOLOv3\\
\hline
Pan et al. \cite{pan2021multi} & 2021 & SL & Fish4Knowledge and Private Datasets & RGB & One & Multi-scale ResNet for Multi-scale Object Detection & SSD Variant\\ 
\hline
Jiang et al. \cite{jiang2021underwater} & 2021 & SL & UODD Dataset & RGB & One & Image Enhancement Framework WaterNet & YOLOv3 Variant\\
\hline
Yu et al. \cite{yu2021real} & 2021 & SL & Private Dataset & Sonar & One & Deep Learning & Transformer–YOLOv5\\
\hline
Yeh et al. \cite{yeh2021lightweight} & 2021 & SL & Private Dataset & RGB & One & Joint Color Conversion and Object Detection & YOLOv3 Variant\\
\hline
Chen et al. \cite{chen2022swipenet} & 2022 & SL & URPC2017 and URPC2018 Datasets & RGB & One & Curriculum Multi-class Adaboost Algorithm to Handle Noisy Data & SWIPENET (SSD Variant)\\
\hline
Alla et al. \cite{alla2022vision} & 2022 & SL & Private Dataset & RGB & One & Image Enhancement Techniques & YOLOV4\\  
\hline
Zhang et al. \cite{zhang2022object} & 2022 & SL & URPC2017 Dataset & RGB & One & - & YOLO Variant\\
\hline
Cai et al. \cite{cai2022underwater} & 2022 & SL & URPC2021 Dataset & RGB & One & Collaborative Weakly Supervision & YOLOv5\\
\hline
Jia et al. \cite{jia2022underwater} & 2022 & SL & URPC Dataset & RGB & One & - & EfficientDet Variant (SSD Variant)\\
\hline
Wang et al. \cite{wang2022underwater} & 2022 & SL & URPC2020 Dataset & RGB & Two & Joint Image Reconstruction and Object Detection & Faster RCNN Variant\\
\hline
Liang et al. \cite{liang2022excavating} & 2022 & SL & UTDAC2020 Dataset & RGB & Two & Attention Module to Capture RoI-level Relation & Faster RCNN Variant\\
\hline
Chen et al. \cite{chen2023htdet} & 2023 & SL & URPC2018 Dataset & RGB & One & - & Hybrid Transformer\\
\hline
Song et al. \cite{song2023boosting} & 2023 & SL & URPC2020 and Brackish Datasets & RGB & Two & Boosting Re-weighting for Hard Example Mining & Boosting RCNN\\
\hline
Dai et al. \cite{dai2023edge} & 2023 & SL & UTDAC2020, Brackish, and TrashCan Datasets & RGB & Two & Edge-guided Attention Module to Capture Boundary Information & Faster RCNN Variant\\
\hline
Fu et al. \cite{fu2023learning} & 2023 & SL & URPC2020 and UODD Datasets & RGB & Two & Incorporate Transferable Priors to Remove Degradation & Cascade RCNN Variant\\
\hline
Zhou et al. \cite{zhou2024amsp} & 2024 & SL & DUO and RUOD Datasets & RGB & One & AMSP-VConv to Handle Noise and Degradation & AMSP-UOD (SSD Variant)\\
\hline
Guo et al. \cite{guo2024lightweight} & 2024 & SL & RUOD, UTDAC2020, and URPC2022 Datasets & RGB & One & Backbone Improvements for Real-time Inference & YOLOv8\\
\hline
Gao et al. \cite{gao2024pe} & 2024 & SL & UTDAC and RUOD Datasets & RGB & One & Path-augmented Transformer to Enhance Small Object Detection& Transformer Variant\\
\hline
Ge et al. \cite{ge2024advanced} & 2024 & SL & UATD Dataset & Sonar & One & - & YOLOv7 Variant\\
\hline
Dai et al. \cite{dai2024gated} & 2024 & SL & DUO, Brackish, TrashCan, and WPBB Datasets & RGB & One & Gated Cross-domain Collaborative Network to Address Poor Visibility and Low Contrast & GCCNet (SSD Variant)\\
\hline
Wang et al. \cite{wang2024dual} & 2024 & SL & DUO, UODD, RUOD, and UDD Datasets & RGB & One & Joint Image Enhancement and Object Detection & DJLNet (SSD Variant)\\
\bottomrule
\end{tabular}
\end{center}
\end{table*}

\begin{figure*}[tb]
\centering
\includegraphics[width=18cm]{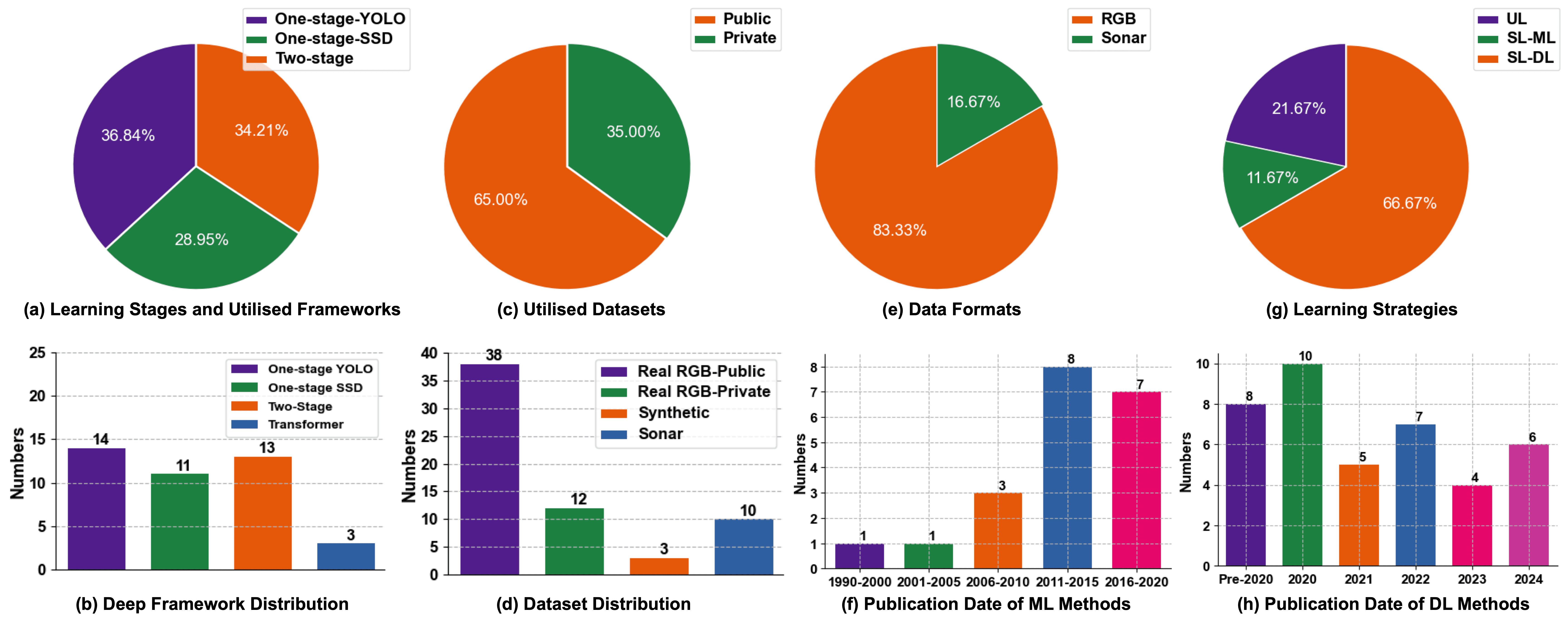}
\caption{A statistic analysis of AI-based UOD methods, including (a, b) utilised frameworks and learning stages, (c, d, e) utilised datasets, (g) learning strategies, and (f, h) publication dates.}
\label{fig:statistics}
\end{figure*}
\begin{figure*}[htb]
\centering
\includegraphics[width=16.5cm]{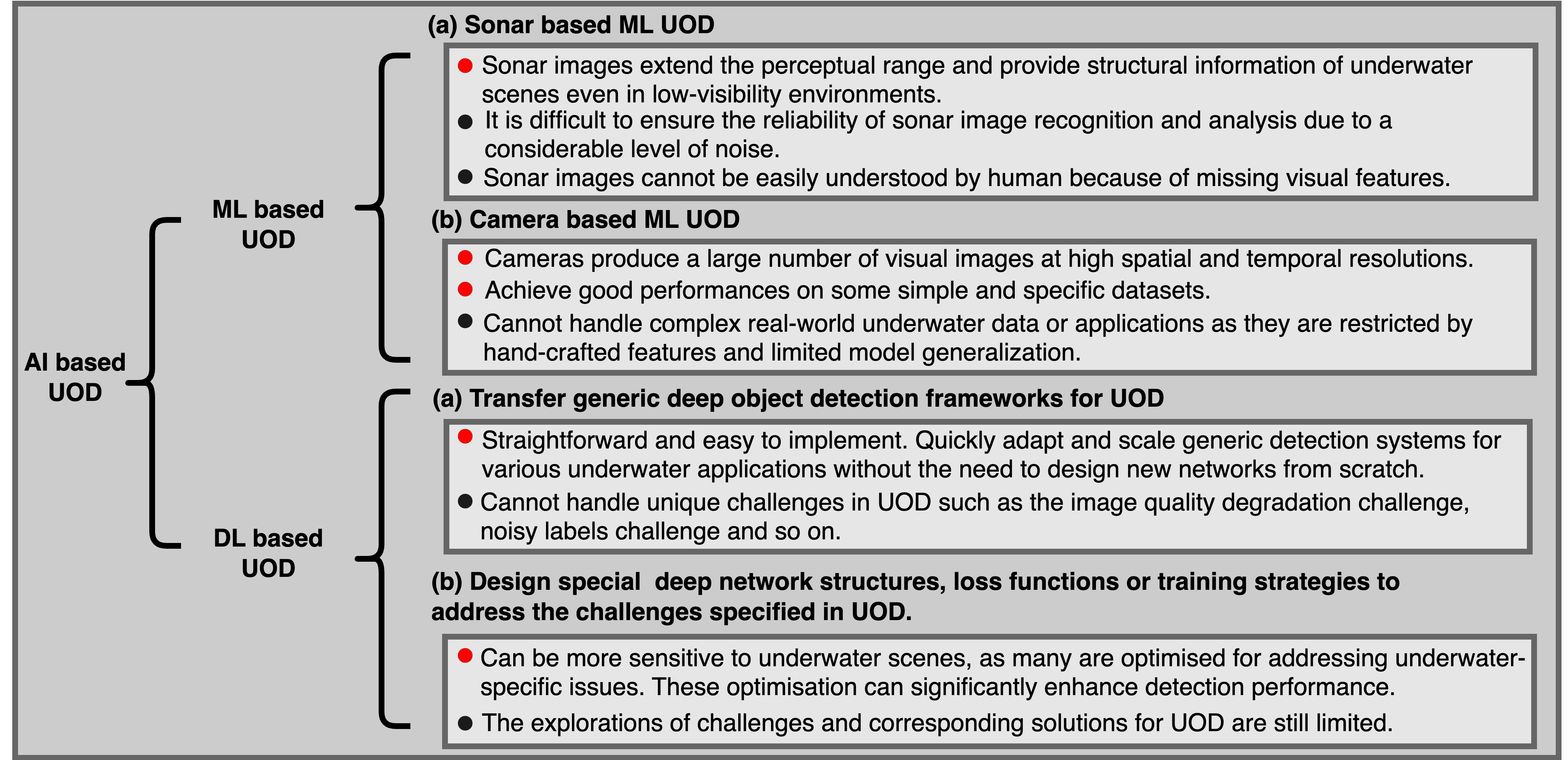}
\caption{The categorisation and summary of previous research in AI-based UOD, highlighting the advantages (\textcolor{red}{red}) and disadvantages of each type.}
\label{fig:AIUOD_summary}
\end{figure*}
\subsection{DL-based Underwater Object Detection}

In deep learning-based UOD, two primary research directions have emerged: transferring generic object detection frameworks for UOD and designing specialised frameworks tailored to underwater environments.

\subsubsection{Transferring Generic Object Detection Frameworks.} A prominent approach in deep learning-based UOD involves directly transferring generic deep detection backbones and frameworks \cite{ren2016faster,liu2016ssd,redmon2016you} to the underwater object detection task. These frameworks are generally categorised into two-stage detectors and one-stage detectors.

\textit{Two-stage Generic Detectors for UOD.} Several studies have leveraged two-stage detectors to extract powerful deep features for UOD. For example, Li et al. \cite{li2015fast} initially applied the generic Fast-RCNN \cite{girshick2015fast} framework for fish species detection and later adopted the Faster-RCNN \cite{ren2016faster} framework to accelerate the detection process in \cite{li2017deep}. In 2018, Ji et al. \cite{ji2018design} introduced an RFCN variant for marine organism detection, while Lee et al. \cite{lee2018deep} and Karimanzira et al. \cite{karimanzira2020object} used Faster RCNN for underwater object detection. Two-stage deep detectors generally offer high localisation and recognition accuracy, however, they fall short in achieving real-time detection due to their complex, coarse-to-fine processing paradigm.

\textit{One-stage Generic Detectors for UOD.} To achieve real-time detection, many generic one-stage detectors have been applied in underwater object detection, particularly SSD variants \cite{liu2016ssd} and YOLO variants \cite{redmon2016you}. Zhang et al. \cite{zhang2020research} utilised the generic SSD framework for underwater object detection. They also applied three underwater image enhancement algorithms to enhance the quality of underwater images and subsequently examined the correlations between image enhancement and object detection tasks. 

YOLO \cite{redmon2016you} and its variants \cite{yang2021research, shkurti2017underwater,cai2022underwater,gavsparovic2022deep} are also among the most frequently used generic detectors for UOD. Yang et al. \cite{yang2021research} utilised the real-time YOLOv3 \cite{farhadi2018yolov3} framework to detect underwater objects in color images. In contrast, Sung et al. \cite{sung2020realistic} and Zhang et al. \cite{zhang2022object} applied the YOLO framework or its variants to underwater object detection. Pedersenet al. \cite{pedersen2019detection} used both YOLOv2 and YOLOv3 for detecting marine animals in underwater scenes with varying visibility. Additionally, YOLOV4 \cite{alla2022vision} and YOLOv5 \cite{cai2022underwater} have also been employed for underwater object detection.

\subsubsection{Specially Designed Underwater Object Detection Frameworks}
Another key research direction in DL-based UOD is the development of novel frameworks or algorithms specially designed to tackle the unique challenges of UOD. Interestingly, we observe that most of these frameworks are one-stage deep detectors, as they can effectively balance accuracy and speed. We categorise these one-stage detectors into SSD variants, YOLO variants, and transformer variants. 

\textit{SSD Variants.} Over the past few years, many UOD frameworks have developed based on the SSD architecture. Zhang et al. \cite{zhang2018single} enhanced the SSD framework by incorporating multi-scale and context features to improve multi-scale object detection in complex underwater environments. Similarly, Zhang et al. \cite{zhang2020mffssd} introduced the MFFSSD framework, which integrates an attention module and a multi-scale feature fusion module into SSD to enhance multi-scale underwater object detection. To tackle the issue of noisy data, Chen et al. \cite{chen2022swipenet} proposed a easy-to-hard learning algorithm, called Curriculum Multi-class Adaboost (CMA), to train deep detection networks. Lin et al. \cite{lin2020roimix} introduced RoIMix, a data augmentation technique that enhances interactions between images and mixing region proposals from multiple images. To achieve real-time underwater object detection, Pan et al. \cite{pan2021multi} proposed a lightweight multi-scale ResNet tailored for underwater environments. 

\textit{YOLO Variants.} Many researchers have focused on enhancing the YOLO framework for UOD. Wang et al. \cite{wang2020yolo} introduced YOLO-Nano-Underwater, a deep detector designed to reduce inference time. Given that deep detectors trained on limited underwater data often suffer from severe domain shift, Liu et al. \cite{liu2020towards} integrated a domain invariant module (DIM) and an invariant risk minimisation (IRM) penalty term into YOLOv3 framework to contruct a detector that performs consistently across various underwater domains. Meanwhile, Jiang et al. \cite{jiang2021underwater} introduced a channel sharpening attention module (CSAM) to enhance the fusion of feature maps with the input image in the YOLOv3 framework. This strategy improves the accuracy of multi-scale object detection, particularly for small and medium-sized objects.

\textit{Transformer Variants.} Several studies have explored the use of transformers in underwater object detection. Yu et al. \cite{yu2021real} integrated a transformer module with the YOLOv5s framework to construct a novel model called TR–YOLOv5s. The transformer module, which includes a self-attention mechanism, allows TR–YOLOv5s to focus more on objects rather than backgrounds. While transformers excel at modeling long-range relationships compared to CNNs, they require more training data and exhibit higher computational complexity. To leverage the strengths of both transformers and CNNs, Chen et al. \cite{chen2023htdet} developed a hybrid transformer network that integrates transformer and CNN modules. This hybrid transformer network captures global contextual information and outperforms both standalone CNNs and transformers.

\subsection{The Summary of AI-based UOD Methods}
Fig. \ref{fig:AIUOD_summary} provides a summary of the strengths and weaknesses of each type of AI-based UOD methods.
\begin{figure*}[htb]
\centering
\includegraphics[width=16cm]{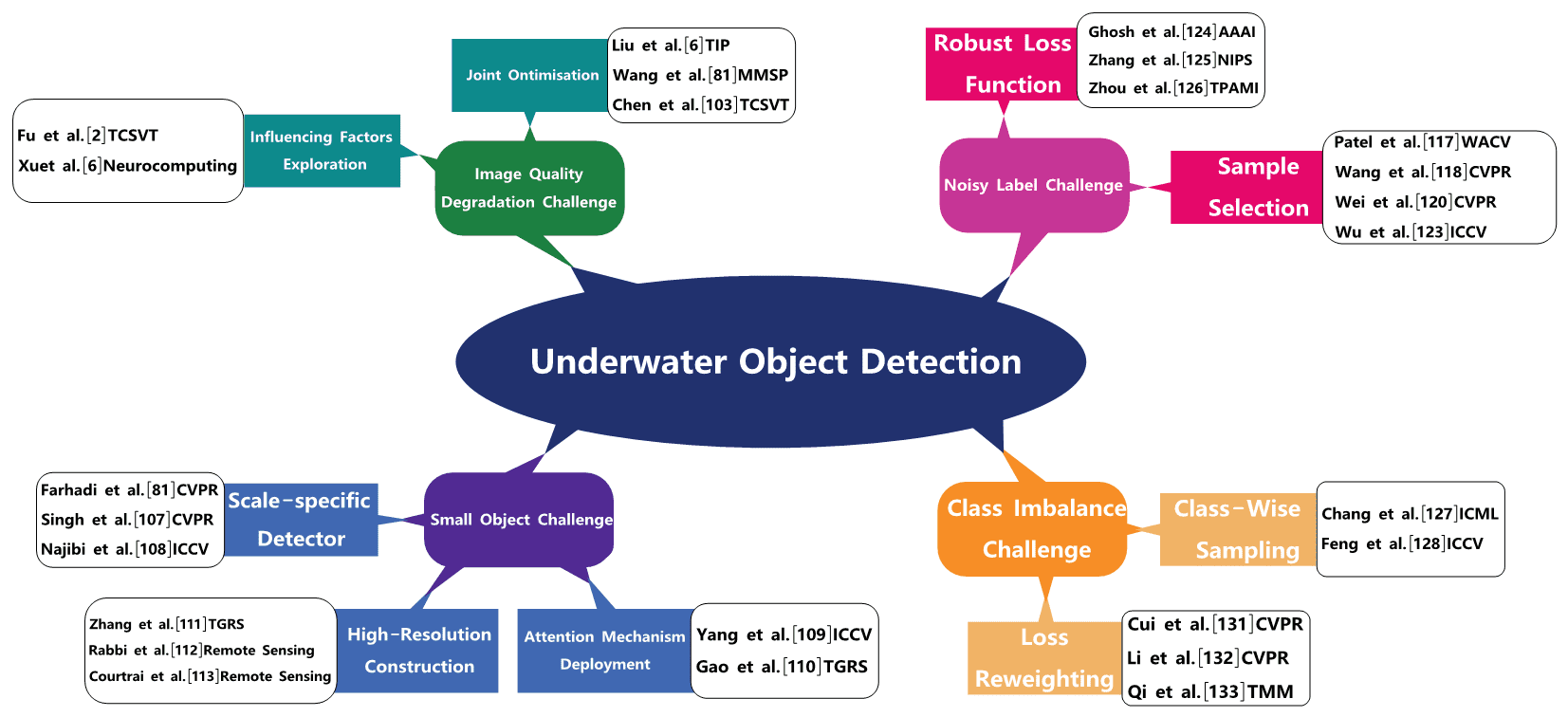}
\caption{Challenges and possible solutions for underwater object detection.}
\label{fig:challenge_solution}
\end{figure*}

\subsubsection{Comparisons of ML UOD and DL UOD Methods}
To understand the strengths and weaknesses of traditional ML and DL UOD methods, we categorise them based on four aspects: learning strategies, applicable datasets, utilised features or frameworks, and learning stages, a statistic analysis from different perspectives is presented in Fig. \ref{fig:statistics}. 

As illustrated in Tables \ref{tab:reviewmachine} and \ref{tab:reviewdeep}, traditional machine learning-based UOD methods utilise supervised or unsupervised learning on relatively small-scale datasets, while deep learning methods typically exploit supervised learning on large-scale datasets. Traditional machine learning approaches generally involve a two-stage learning process: feature extraction and detector selection or design. In contrast, deep learning approaches commonly employ a one-stage, end-to-end learning pipeline. Traditional machine learning methods have recently fallen behind advanced deep learning techniques. This is because they are limited to a narrow set of hand-crafted features and cannot fully leverage the potential of large-scale datasets. On the other hand, deep learning models require large amounts of training data but offer clear advantages over traditional methods in terms of detection accuracy.

\subsubsection{Comparison of Transferred Generic Detectors and Specially Designed Underwater Detectors}

Transferring generic object detection frameworks for UOD is straightforward and easy to implement. This enables researchers to quickly adapt and scale detection systems for various underwater applications without the need to design new networks from scratch. Moreover, generic detection models are typically pre-trained on large, diverse datasets, offering a strong foundation of learned features that can be fine-tuned for underwater environments. 

However, generic object detectors trained on terrestrial objects may not adequately capture the unique characteristics of underwater environments, such as specific colors, textures, and shapes of marine objects, leading to reduced detection performance. In contrast, specially designed underwater detectors can be more sensitive to underwater image degradation, as many are optimised for underwater-specific issues like color distortions, haze, blurring, and low visibility. These optimisation can significantly enhance detection performance, as the models are tailored to address these challenges.
\begin{table*}[t]
\label{tab:surveys}
\begin{center}
\caption{A summary of existing underwater object detection datasets.}
\label{tab:datasets}
\begin{tabular}{p{2.5cm}p{1.2cm}p{1.5cm}p{1.5cm}p{0.6cm}p{1cm}p{1.0cm}p{1cm}p{2cm}}
\toprule
Datasets   & Train & Test/Val & Total & Cls & Anno. & Object & Year & Download Link\\
\hline
F4K-Species \cite{boom2012supporting}      & - & - & 27,370& 23& Mask & -      & 2012 & -\\
F4K-Trajectory \cite{beyan2013detecting} & - & - & 93 videos & - & BBox & -  & 2013 & -\\
F4K-Complex \cite{kavasidis2014innovative} & - & - & 14 videos & - & BBox & -  & 2014 & -\\
Brackish \cite{pedersen2019detection} & 11,739 & 1,468/1467 & 14,674& 6 & BBOX & 35,565 & 2019 & \href{https://www.kaggle.com/datasets/aalborguniversity/brackish-dataset}{Link}\\
URPC2017   & 17,655 & 985   & 18,640& 3 & BBOX & -      & 2017 & \href{https://pan.baidu.com/s/1puJ0BRvoTTaPjk4_KzhEKw}{Link (pwd:0hct)}\\
URPC2018   & 2,901  & 800   & 3,701 & 4 & BBOX & 22,688 & 2018 & \href{https://pan.baidu.com/s/1puJ0BRvoTTaPjk4_KzhEKw}{Link (pwd:0hct)}\\
URPC2019   & 4,757  & 1,029 & 5,786 & 4 & BBOX & 36,100 & 2019 & \href{https://drive.google.com/file/d/1n8Rpgx3xF84HO6PXpfPrRtTMtVSuOaBs/view}{Link}\\
URPC2020   & 6,575  & 2,400 & 8,975 & 4 & BBOX & 46,287 & 2020 & \href{https://github.com/xiaoDetection/Learning-Heavily-Degraed-Prior}{Link}\\
URPC2021   & 7,478  & 1,200 & 8,678 & 4 & BBOX & 54,238 & 2021 & -\\
UDD \cite{liu2021new}     & 1,827  & 400   & 2,227 & 3 & BBOX & 15,022 & 2021 & \href{https://github.com/chongweiliu/UDD_Official}{Link}\\
DUO \cite{liu2021dataset} & 6,671  & 1,111 & 7,782 & 4 & BBOX & 74,515 & 2021 & \href{https://github.com/chongweiliu/DUO}{Link}\\
UODD \cite{jiang2021underwater} & 2,688  & 506   & 3,194 & 3 & BBOX & 19,212 & 2021 & \href{https://github.com/LehiChiang/Underwater-object-detection-dataset}{Link}\\
RUOD \cite{fu2023rethinking}    & 9,800  & 4,200 & 14,000&10 & BBOX & 74,903 & 2022 & \href{https://github.com/dlut-dimt/RUOD}{Link}\\
\hline
\bottomrule
\end{tabular}
\end{center}
\end{table*}

\begin{figure*}[htb]
\centering
\includegraphics[width=16cm]{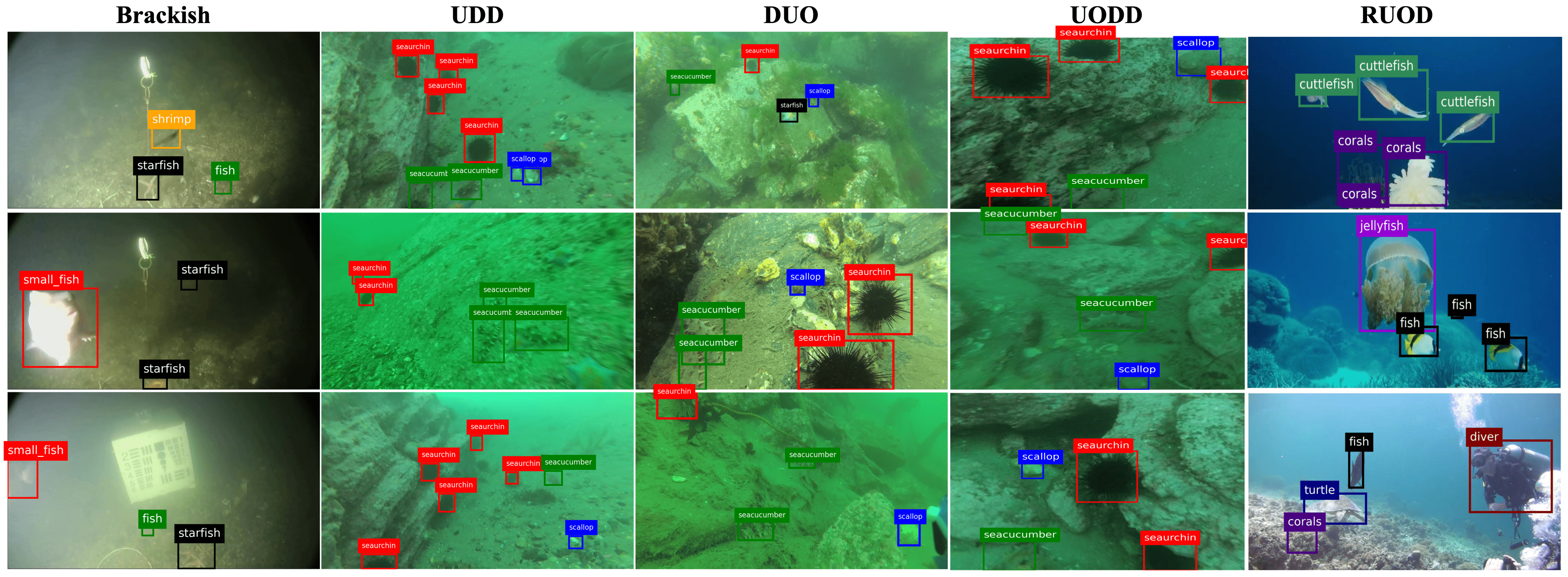}
\caption{Visual comparisons of images and annotations in some representative underwater object detection datasets.}
\label{fig:example_im}
\end{figure*}

\begin{figure*}[htb]
\centering
\includegraphics[width=16cm]{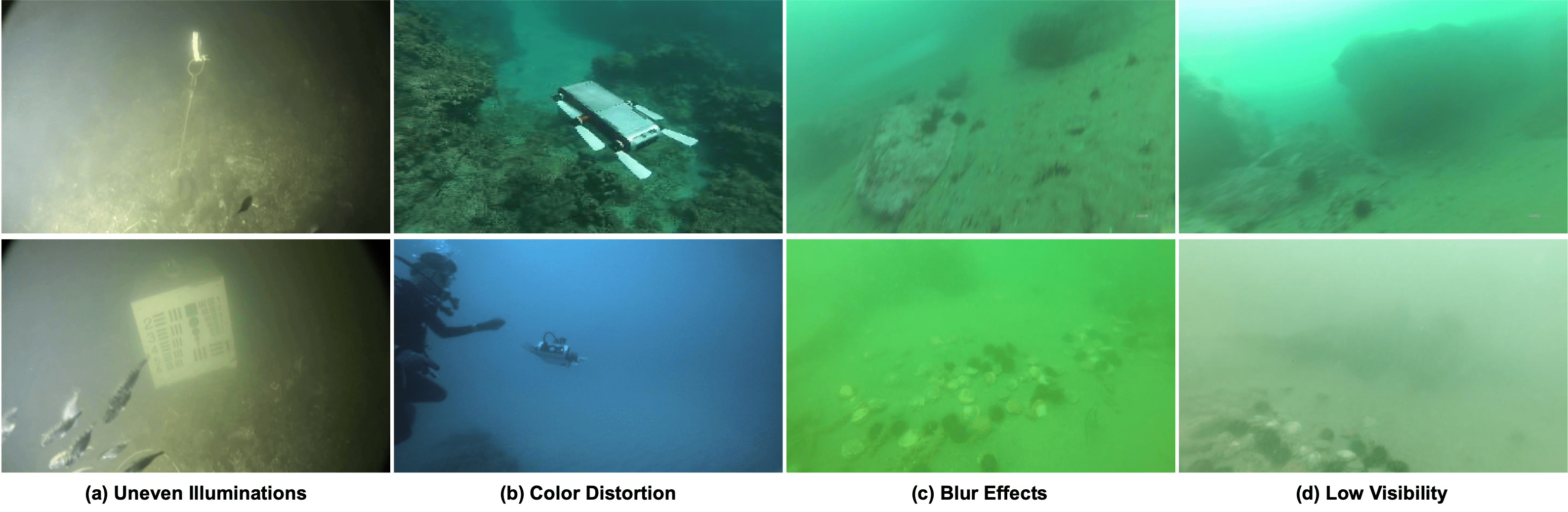}
\caption{Examples of image quality degradation challenges include (a) uneven illuminations, (b) color distortions, and (c) blur effects, and (d) low visibility.}
\label{fig:degradation}
\end{figure*}

\section{Challenges and Future Solutions}
\label{sec:challenge_solution}

Underwater scenes are among the most challenging environments for object detection due to their specific difficulties. To facilitate a clearer understanding of the challenges, we categorise them into four groups: image quality degradation, small object detection, noisy labels, and class imbalance. In this work, we review previous research, summarise existing solutions for each challenge in both UOD and GOD, and suggest potential improvements and research directions, as illustrate in Fig. \ref{fig:challenge_solution}.

\subsection{Image Quality Challenges and Solutions}

Underwater environments face significant image quality degradation due to physical phenomena like light absorption and scattering \cite{chiang2011underwater}, along with the use of artificial lighting, as illustrated in Fig. \ref{fig:degradation}. These physical phenomena significantly degrade image quality, posing huge challenges to UOD. Most previous works \cite{fabbri2018enhancing, chen2019towards} first apply underwater image enhancement techniques to improve image quality before using the enhanced images to train UOD frameworks. While it is generally assumed that UIE can boost UOD performance by enhancing visual quality, several studies \cite{liu2020real, fu2023rethinking} have examined the relationship between UIE and found no strong correlation between image quality and detection accuracy. In some cases, UIE algorithms can even degrade detection performance \cite{xu2023systematic}. 

These findings indicate that separate optimisation of UIE and UOD can result in inconsistent outcomes and suboptimal solutions, as the two tasks have different optimisation objectives: UIE is opitimised using visual quality-based losses, while UOD relies on detection accuracy-based losses. 

\textit{Potential Solutions:} 1) \textit{Exploring Influencing Factors.} Begin by identifying the key factors that contribute to performance drops in UOD, and then develop targeted solutions. For instance, Fu et al. \cite{fu2023rethinking} summarised the challenges posed by image quality degradation into three types: color distortion, haze effect, and light interference. Their research revealed that color distortion and light interference are the primary factors causing performance drops in detectors, while haze effect has a relatively minor impact. Therefore, it is essential to investigate these interfering factors in depth.

2) \textit{Joint Optimisation of UIE and UOD.} Given the semantic gaps between UIE and UOD, enhancing their interactions to bridge these gaps is a promising research direction. For instance, Chen et al. \cite{chen2020perceptual} introduced a detection perceptor that guides the enhancement model to produce images favorable for detection, incorporating an object-focused perceptual loss to jointly supervise the optimisation of the enhancement model. Liu et al. \cite{liu2022twin} introduced a task-aware feedback module into the enhancement pipeline. This module provides detection-related information to the enhancement model, directing it towards generating detection-favoring images. Additionally, Wang et al. \cite{wang2022underwater} proposed a joint framework for simultaneous underwater object detection and image reconstruction, the joint optimisation enables the shared network backbone to better generalise and adapt to various underwater scenes.

\subsection{Small Object Detection Challenges and Solutions }

Many objects in underwater images occupy only a tiny fraction of the image. For instance, objects in the DUO dataset \cite{liu2021dataset} typically cover only 0.3\% to 1.5\% of the image area. Even advanced deep detectors often struggle to identify these small objects effectively. Several approaches \cite{pan2021multi, dai2021attentional, zhang2021lightweight} have been developed to tackle the challenge of small object detection in underwater scenes. Here, we review advanced techniques in both GOD and UOD to facilitate small object detection.

\textit{Potential Solutions:} 1) \textit{Scale-specific Detector Design.} Since features at different layers are sensitive to objects of special scales, many approaches have developed multi-branch architectures to construct scale-specific detectors. Farhadi et al. \cite{farhadi2018yolov3} designed parallel branches to detect multi-scale objects, with one branch focused on constructing high-resolution feature maps to capture small objects. Singh et al. \cite{singh2018analysis} introduced a novel training paradigm called Scale Normalization for Image Pyramids (SNIP), which trains deep detectors using only objects that fall within the desired scale range, effectively addressing small object at the most relevant scales. Similarly, Najibi et al. \cite{najibi2019autofocus} proposed a coarse-to-fine pipeline that processes small objects at the most reasonable scale.

2) \textit{Attention Mechanism Deployment.} The attention mechanism, designed to emphasise regions of interests, offers another notable solution for small object detection. Yang et al. \cite{yang2019scrdet} introduced pixel and channel attention modules to enhance regions containing small objects. Gao et al. \cite{gao2023global} developed a local attention pyramid module to enhance the feature representation of small objects while suppressing backgrounds and noise in shallower feature maps. These attention-based solutions offer high flexibility and can be plugged into almost any deep detection architectures, however, they often come with substantial computational overhead due to the correlation operations involved.

3) \textit{High-resolution Image Reconstruction.} This technique seeks to improve the details of small objects by increasing image resolution. Zhang et al. \cite{zhang2023superyolo} incorporated a super-resolution branch within the detection framework to learn high-resolution feature representations for small objects. Both Rabbi et al. \cite{rabbi2020small} and Courtrai et al. \cite{bashir2021small} employed GANs to super-resolve low-resolution remote sensing images, enhancing edge details and minimising high-frequency information loss during reconstruction. While high-resolution image reconstruction enriches the feature representation of small objects by amplifying image resolution, it may also introduce spurious textures and artifacts, which can negatively impact detection performance. 

\subsection{Noisy Label Challenge and Solutions}

In real-world datasets, the proportion of noisy/incorrect labels has been reported to range from 8.0\% to 38.5\% \cite{song2022learning,song2019selfie}. The noisy label problem is particularly pronounced in underwater object detection datasets, where images often suffer from high levels of blur and low visibility, making accurate data annotation more challenging. Here, we review existing solutions for handling noisy labels in the deep learning community and summarise key approaches that may be well-suited for UOD.

\textit{Potential Solutions:} 1) \textit{Sample Selection.} Sample selection \cite{wang2018iterative, patel2023adaptive} is a widely used strategy for filtering out clean samples from noisy datasets. Both theoretical \cite{wang2021proselflc} and empirical studies \cite{ma2023ctw} demonstrate that deep learning models first learn simple, generalised patterns before gradually overfitting to noisy patterns. Based on this expertise, quite a few works \cite{wei2020combating, yao2020searching} select small-loss training samples as clean examples in the early stages of training. More advanced approaches have utilised multi-network learning \cite{li2017learning} and multi-round learning \cite{wu2021ngc} to iterative refine the selection of clean samples. These methods are well-motivated, however, they can suffer from accumulated errors from incorrect sample selection, and finding a robust strategy for noisy sample selection remains challenging.

2) \textit{Robust Loss Function.} Several noise-robust loss functions have been developed to train deep learning models on datasets with noisy labels. Ghosh et al. \cite{ghosh2017robust} introduce a noise-tolerant loss function for multi-class classification, which performs well under symmetric noise but requires knowledge of the noise rate. Zhang et al. \cite{zhang2018generalized} proposed a generalised cross-entropy (GCE) loss function that combines the benefits of both mean absolute error and categorical cross-entropy losses. Zhou et al. \cite{zhou2023asymmetric} introduced an asymmetric loss function that satisfy the Bayes-optimal condition, making them robust to noisy labels. While these loss functions can enhance the generalisation ability of deep models in certain situations, they often depend on special conditions, such as a known noise rate.

\subsection{Class Imbalance Challenge and Solutions}

\begin{figure*}[htb]
\centering
\includegraphics[width=18cm]{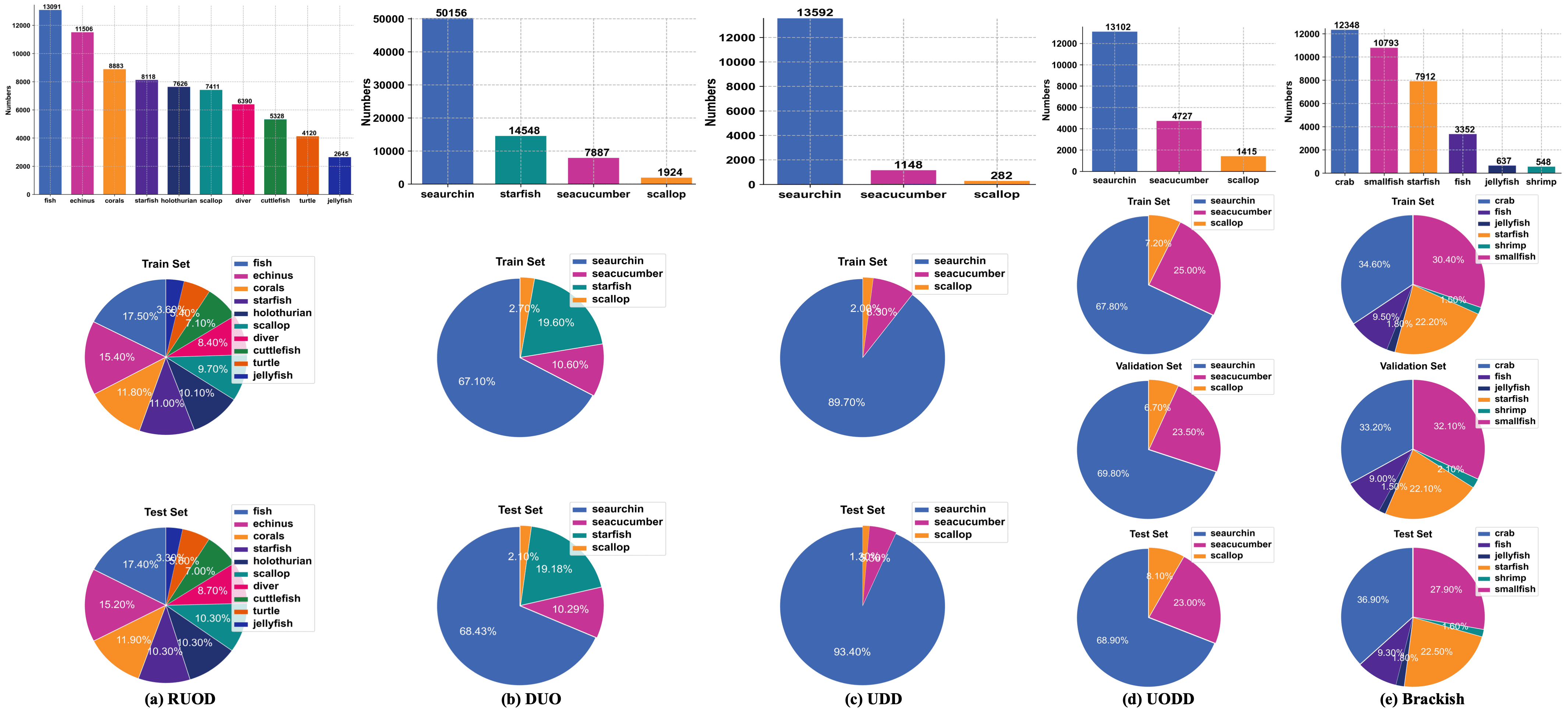}
\caption{Data distributions in the representative underwater object detection datasets show that class imbalance is a widespread issue.}
\label{fig:data_distribution}
\end{figure*}

Long-tail distribution is frequently observed in underwater object detection datasets, as illustrate in Fig. \ref{fig:data_distribution}. Here, we present two strategies to tackle class imbalance issues in UOD.

1) \textit{Class-Aware Sampling.} Class-aware sampling seeks to address class imbalance by adjusting the class distribution through class-wise resampling during training. For instance, Chang et al. \cite{chang2021image} identified limitations in image-level resampling strategies for long-tailed detection and proposed an object-centric sampling method. Feng et al. \cite{feng2021exploring} introduced a Memory-Augmented Feature Sampling (MFS) module designed to over-sample features from underrepresented classes. While class-aware sampling helps generate a more balanced data distribution, it may also introduce large amounts of duplicated samples, which can slow down training and increase the risk of overfitting.

2) \textit{Loss Reweighing.} Loss reweighing~\cite{ren2018learning} aims to enhance the learning of minority classes by increasing their weights. For instance, Mahajan et al. \cite{mahajan2018exploring} reweighted classes using the square root of their frequency, while Cui et al. \cite{cui2019class} employed the effective number of samples rather than raw class frequency to balance the classes. Li et. al \cite{li2020overcoming} presented a Balanced Group Softmax (BAGS) module, which optimises the classification head in the detection framework to ensure adequate training for all classes. Observing that learning bias arises not only from class distribution disparities but also from differences in sample diversity, Qi et al. \cite{qi2023balanced} proposed the BAlanced CLassification (BACL) framework to address imbalances stemming from both class distribution and sample diversity. All these works provide insights for addressing class imbalance in underwater object detection and require further investigation to validate their effectiveness.

\section{Datasets, Evaluation Metrics, and Detection Diagnosis Tools}
\label{sec:datasets}

In this section, we first provide a comprehensive overview of previous UOD datasets, followed by a discussion of commonly used evaluation metrics in UOD. Finally, we introduce two valuable detection diagnosis tools that help identify strengths and weaknesses of each underwater detector.

\subsection{Datasets}

Several underwater object detection datasets have been proposed over the past decade. In this subsection, we review some of the most representative datasets, including the Fish4K \cite{fisher2016fish4knowledge}, Brackish \cite{pedersen2019detection}, the URPC series, UDD \cite{liu2021new}, DUO \cite{liu2021dataset}, UODD \cite{jiang2021underwater}, and RUOD \cite{fu2023rethinking}. A summary of these datasets is presented in Table \ref{tab:datasets} and Fig. \ref{fig:data_distribution}.

\textit{Fish4K Dataset \cite{fisher2016fish4knowledge}.} The Fish4K dataset is one of the most well-known early-stage underwater datasets for fish detection and species classification. It offers a large collection of videos and images captured by 10 cameras between 2010 and 2013. The dataset includes a diverse range of moving organisms, such as swimming fish, sea anemones, algae, and aquatic plants. Numerous detection and classification algorithms \cite{vasamsetti2018automatic, pan2021multi} have ever been developed and validated on various subsets of Fish4K, such as F4K-Complex \cite{kavasidis2014innovative}, F4K-Species \cite{boom2012supporting}, and F4K-Trajectories \cite{beyan2013detecting}. However, the download links for this dataset and its subsets are no longer available.

\textit{Brackish dataset \cite{pedersen2019detection}.} The Brackish dataset is Kaggle competition dataset for fish detection, collected in brackish water with varying levels of visibility. The dataset consists of images extracted from 89 videos, annotated with bounding boxes. Fish are categorised into six broad groups: fish, small fish, crab, shrimp, jellyfish, and starfish. The dataset is divided into training, validation, and testing sets, following an 80/10/10 split. In total, it contains 14,674 underwater images with 35,565 annotated instances.

\textit{URPC Series Datasets.} The Underwater Robot Professional Contest (URPC)\footnote{\url{http://www.urpc.org.cn/index.html}} has been held annually since 2017, providing large-scale underwater object detection datasets for competition purposes each year. To date, URPC has released five datasets (URPC 2017-2021). The underwater images are captured near Zhangzi Island, Dalian, China, and all URPC datasets include bounding box annotations. Among them, URPC2017 features three object categories (scallop, seacucumber, and seaurchin), while later datasets include four categories (scallop, starfish, seacucumber, and seaurchin). However, the download links for these dataset are no longer available after the competition. In Table \ref{tab:datasets}, we provide the download links sourced from unofficial researchers.

\textit{UDD Dataset \cite{liu2021new}.} The UDD is a benchmark designed for underwater object detection tasks, featuring 15,022 instances with bounding box annotations. The images are collected from an open-sea farm in Zhangzi Island, Dalian, China. The dataset comprises 2,227 images, divided into a training set of 1,827 images and a testing set of 400 images. It includes three object categories: scallop, seacucumber, and seaurchin.

\textit{DUO Dataset \cite{liu2021dataset}.} The DUO dataset is generated by re-annotating the URPC series (2017-2020) and UDD datasets, as many of these datasets do not offer testing sets and their images suffer from poor annotation quality. The DUO dataset includes four object categories-scallop, starfish, seacucumber, and seaurchin-featuring improved and more accurate annotations. The dataset comprises 6,671 images for training and 1,111 images for testing. 

\textit{UODD Dataset \cite{jiang2021underwater}.} The UODD dataset is another specialised dataset for underwater object detection, also collected from Zhangzi Island, Dalian, China. It includes 3,194 underwater images with 19,212 annotated instances, divded into 2,688 training images and 506 testing images. The dataset features three object categories: scallop, seacucumber, and seaurchin.

\textit{RUOD Dataset \cite{fu2023rethinking}.} The RUOD dataset is created to include a wide range of object categories and visual variations. RUOD features underwater images collected from various locations across the internet, capturing diverse visual effects such as haze, color distortion, and varying light conditions. RUOD encompasses ten underwater object categories (fish, seaurchin, coral, starfish, seacucumber, scallop, diver, cuttlefish, turtle and jellyfish), 14,000 images, and 74,903 annotated objects. Fig. \ref{fig:example_im} displays examples from RUOD alongside those from previous datasets, illustrating that RUOD features a greater diversity of object categories and visual variations.

\begin{table*}[t]
\begin{center}
\caption{The quantitative performance of representative detection frameworks on the RUOD and DUO datasets.}
\renewcommand{\arraystretch}{1} 
\begin{tabular*}{0.88\linewidth}{ccllrrcccccc}
\toprule
\textbf{Datasets} & \textbf{Types} & \textbf{Methods}    & \textbf{Backbone}         & \textbf{Params}  & \textbf{FLOPs}   & \textbf{AP}   & \textbf{AP$_{50}$} & \textbf{AP$_{75}$}  &  \textbf{AP$_{s}$}  & \textbf{AP$_{m}$} & \textbf{AP$_{l}$}\\
\midrule
\multirow{15}{0.5cm}{\rotatebox{90}{\textbf{RUOD}}} & \multirow{8}{0.5cm}{\rotatebox{90}{\textbf{Generic}}} & RetinaNet \cite{lin2017focal}      & ResNet101     &  55.51M & 273.41G & 52.8	& 81.8 & 57.9 &	14.6 & 39.9 & 58.3\\
& & GuidedAnchor \cite{wang2019region} & ResNet101     &  60.98M & 258.05G & 46.6 & 80.8 & 48.3 & 21.5 & 41.1 & 50.6\\
& & CascadeRCNN \cite{cai2019cascade}  & ResNet101     &  88.17M & 301.06G & 49.8 & 80.5 & 54.2 & 18.7 & 43.1 & 54.3\\
& & RepPoints \cite{yang2019reppoints} & ResNet101     &  55.82M & 256.00G & 53.2 & 82.2 & 60.1 & 28.2 & 44.9 & 57.8\\
& & FoveaBox \cite{kong2020foveabox}   & ResNet101     &  56.68M & 268.29G & 44.8 & 80.2 & 45.2 & 18.0 & 37.5 & 49.1\\
& & ATSS \cite{zhang2020bridging}      & ResNet101     &  51.13M & 267.26G & 54.0 & 80.3 & \textbf{60.2} & 18.0 & 40.0 & 59.5\\
& & DetectoRS \cite{qiao2021detectors} & DResNet50     & 123.23M &  90.05G & 53.3 & \textbf{84.1} & 58.7 & \textbf{30.8} & \textbf{46.6} & 57.8\\
& & GridRCNN \cite{lu2019grid}         & ResNet101     & 129.63M & 365.60G & \textbf{54.2} & 81.6 & 60.0 & 25.6 & 46.1 & \textbf{59.8}\\
\cline{2-12}
& \multirow{7}{0.5cm}{\rotatebox{90}{\textbf{Underwater}}} & RoIMix \cite{lin2020roimix}        & ResNet50      & 68.94M & 91.08G  & 54.6 & 81.3 & 60.3 & \textbf{15.6} & 41.7 & 59.8\\
& & RoIAttn \cite{liang2022excavating} & ResNet50      & 55.23M & 331.39G & 52.9 & 81.7 & 57.3 & 12.2 & 39.0 & 58.3\\
& & BoostRCNN \cite{song2023boosting}  & ResNet50      & 45.95M & 54.71G  & 53.9 & 80.6 & 59.5 & 11.6 & 39.0 & 59.3\\
& & RFTM \cite{fu2023learning}         & ResNet50      & 75.58M & 91.06G  & 53.3 & 80.2 & 57.7 & 11.8 & 39.2 & 59.3\\
& & ERLNet \cite{dai2023edge}          & SiEdgeR50     & 45.95M & 54.71G  & 54.8 & 83.1 & 60.9 & 14.7 & 41.4 & 59.8\\ 
& & GCCNet \cite{dai2024gated}         & SwinFT        & 38.31M & 78.93G  & 56.1 & 83.2 & 60.5 & 11.7 & \textbf{41.9} & 62.1\\
& & DJLNet \cite{wang2024dual}         & ResNet50      & 58.48M & 69.51G  & \textbf{57.5} & \textbf{83.7} & \textbf{62.5} & 15.5 & 41.8 & \textbf{63.1}\\
\bottomrule
\multirow{15}{0.5cm}{\rotatebox{90}{\textbf{DUO}}} & \multirow{8}{0.5cm}{\rotatebox{90}{\textbf{Generic}}} & RetinaNet \cite{lin2017focal}      & ResNet101     &  55.38M & 289.79G & 50.9 & 73.1 & 59.7 & 51.0 & 50.1 & 53.1\\
& & CascadeRCNN \cite{cai2019cascade}  & ResNet101     &  88.15M & 319.49G & 60.6 & 80.9 & 70.5 & 54.4 & 61.4 & 61.2\\
& & RepPoints \cite{yang2019reppoints} & ResNet101     &  55.82M & 256.00G & 59.4 & 80.4 & 70.1 & 55.5 & 59.6 & 60.1\\
& & GridRCNN \cite{lu2019grid}         & ResNet101     & 129.63M & 365.60G & 56.6 & 78.9 & 67.2 & 50.3 & 56.7 & 57.4\\
& & FoveaBox \cite{kong2020foveabox}   & ResNet101     &  55.24M & 286.72G & 53.7 & 78.4 & 63.9 & 55.3 & 54.3 & 54.6\\
& & ATSS \cite{zhang2020bridging}      & ResNet101     &  51.11M & 286.72G & 55.4 & 79.2 & 63.2 & 55.7 & 55.7 & 56.0\\
& & DetectoRS \cite{qiao2021detectors} & DResNet50     & 123.23M &  90.05G & 58.9 & 81.4 & 68.3 & 49.6 & 57.6 & 61.8\\
& & GuidedAnchor \cite{wang2019region}  & ResNet101    &  60.94M & 276.48G & \textbf{61.4} & \textbf{83.8} & \textbf{72.0} & \textbf{58.9} & \textbf{62.4} & \textbf{61.3}\\
\cline{2-12}
& \multirow{7}{0.5cm}{\rotatebox{90}{\textbf{Underwater}}} & RoIMix \cite{lin2020roimix}        & ResNet50  & 68.94M & 91.08G  & 61.0 & 80.0 & 69.7 & 48.0 & 62.5 & 60.2\\ 
& & RoIAttn \cite{liang2022excavating} & ResNet50  & 55.23M & 331.39G & 58.7 & 79.5 & 66.5 & 45.5 & 59.6 & 58.5\\
& & BoostRCNN \cite{song2023boosting}  & ResNet50  & 45.95M & 54.71G  & 60.8 & 80.6 & 69.0 & 48.3 & 62.3 & 59.4\\
& & RFTM \cite{fu2023learning}         & ResNet50  & 75.58M & 91.06G  & 60.1 & 79.4 & 68.1 & 49.0 & 61.1 & 59.5\\ 
& & ERLNet \cite{dai2023edge}          & SiEdgeR50 & 45.95M & 54.71G  & 61.2 & 81.4 & 69.5 & 55.2 & 62.2 & 60.8\\ 
& & GCCNet \cite{dai2024gated}         & SwinFT    & 38.31M & 78.93G  & 61.1 & 81.6 & 67.3 & 52.5 & 63.6 & 59.3\\
& & DJLNet \cite{wang2024dual}         & ResNet50  & 58.48M & 69.51G  & \textbf{65.6} & \textbf{84.2} & \textbf{73.0} & \textbf{55.6} & \textbf{67.4} & \textbf{64.1}\\
\bottomrule
\end{tabular*}
\label{tab:quantitative}
\end{center}
\end{table*}

\begin{figure*}[htb]
\centering
\includegraphics[width=17cm]{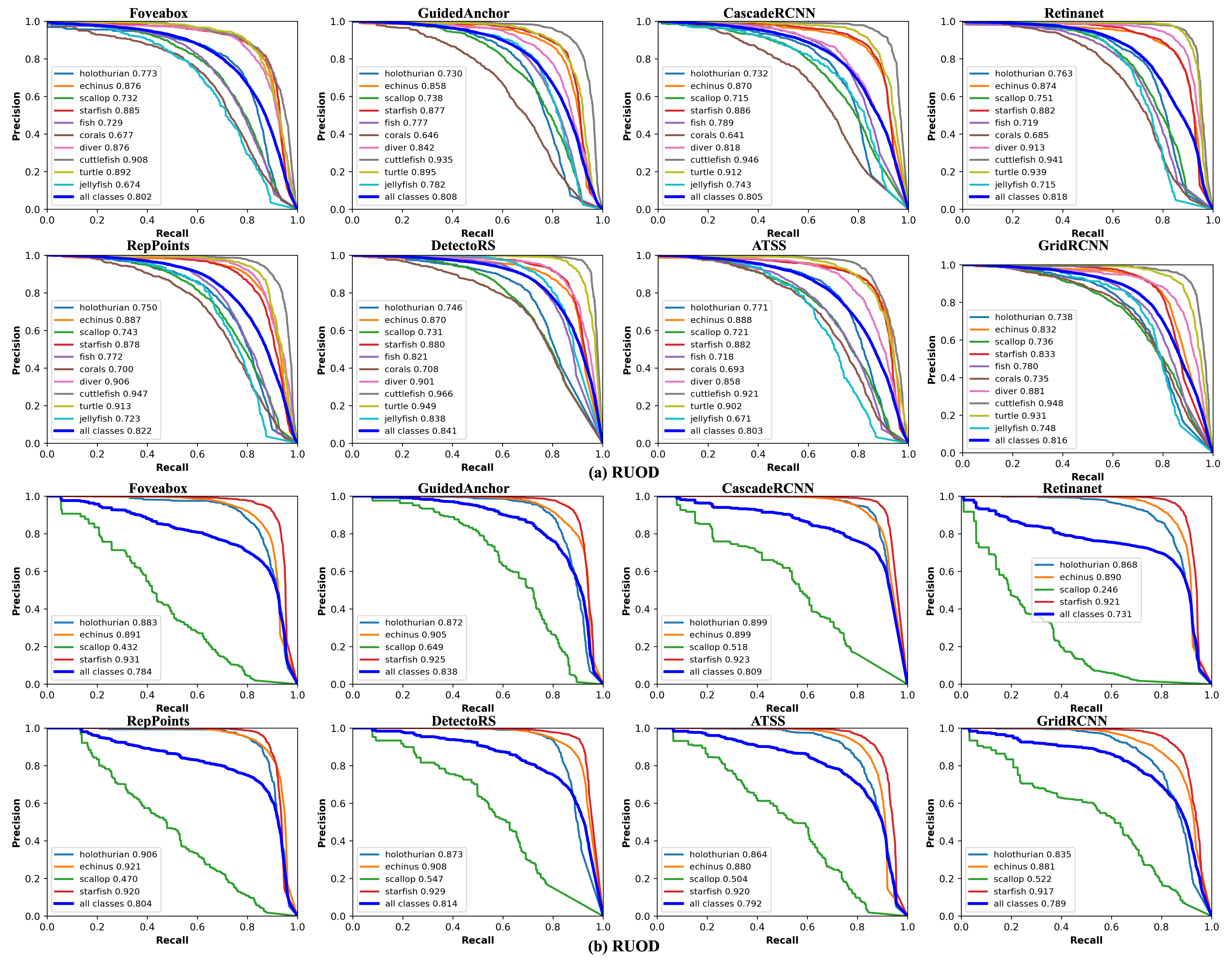}
\caption{Precision-recall curves of representative detection frameworks for each category in the RUOD dataset.}
\label{fig:object_imbalance_RUOD}
\end{figure*}

\subsection{Evaluation Metrics}

Most of evaluation metrics in underwater object detection are adopted from generic object detection. In both fields, evaluation metrics assess the performance of models in terms of both accuracy and efficiency.

\subsubsection{Accuracy Metrics} The accuracy metrics assess the model's accuracy in object detection and include Average Precision (AP) and mean Average Precision (mAP). Average precision (AP) represents the area under the precision-recall curve as shown in Fig. \ref{fig:object_imbalance_RUOD}. It essentially reflects the precision averaged across all recall values from 0 to 1, and can be formulated as: 
\begin{equation}
AP@\alpha=\int_{0}^{1}p(r)dr
\label{eq:AP}
\end{equation}
Here, $p(r)$ represents the precision-recall curve. The notation $AP@\alpha$ indicates that AP is evaluated at a specific IoU threshold $\alpha$. For example, metrics like AP@.50 and AP@.75 refer to the AP calculated at IoU thresholds of 0.5 and 0.75, respectively. For each class, an individual AP can be calculated. The mean Average Precision (mAP) is then obtained by averaging the AP values over all ($n$) classes.

\begin{equation}
mAP@\alpha=\frac{1}{n}\sum_{i=1}^n AP_i
\label{eq:IOU}
\end{equation}
To avoid bias that might arise from using a single IoU threshold, the COCO \cite{lin2014microsoft} mAP evaluator computes mAP across 10 IoU thresholds from 0.5 to 0.95 in increments of 0.05 (mAP@[0.5:0.05:0.95]). Additionally, COCO introduces metrics to assess detection performance across different object scales: $AP_{s}$ for small objects, $AP_{m}$ for medium objects, and $AP_{l}$ for large objects.

\subsubsection{Efficiency Metrics} The efficiency metrics measure the detection model's performance in terms of computational complexity, model size, and real-time inference speed. These metrics include FLOPs, Params, and FPS.

FLOPs measures computational complexity by counting the number of floating-point operations the model required. Models with fewer FLOPs are more computationally efficient. Params refer to the number of learnable weights in a model. Models with fewer parameters are more lightweight, which is better suited for deployment on resource-constrained devices. FPS measures the inference speed of the model, indicating how many frames the model can process per second.

\subsection{Detection Analysis Tools}

Object detectors can fail due to various issues, such as misclassification, localisation errors, or missed detection. High-level metrics like mAP offer limited insight into the root causes of underperformance. Tools that provide a quantitative breakdown of error types offer researchers a deeper understanding of where and why detectors fail, helping identify areas for improvements. Here, we introduce two practical and easy-to-use detection tools: Diagnosis \cite{hoiem2012diagnosing} and TIDE \cite{bolya2020tide}, each offering distinct yet complementary advantages. 

\begin{figure*}[htb]
\centering
\includegraphics[width=18cm]{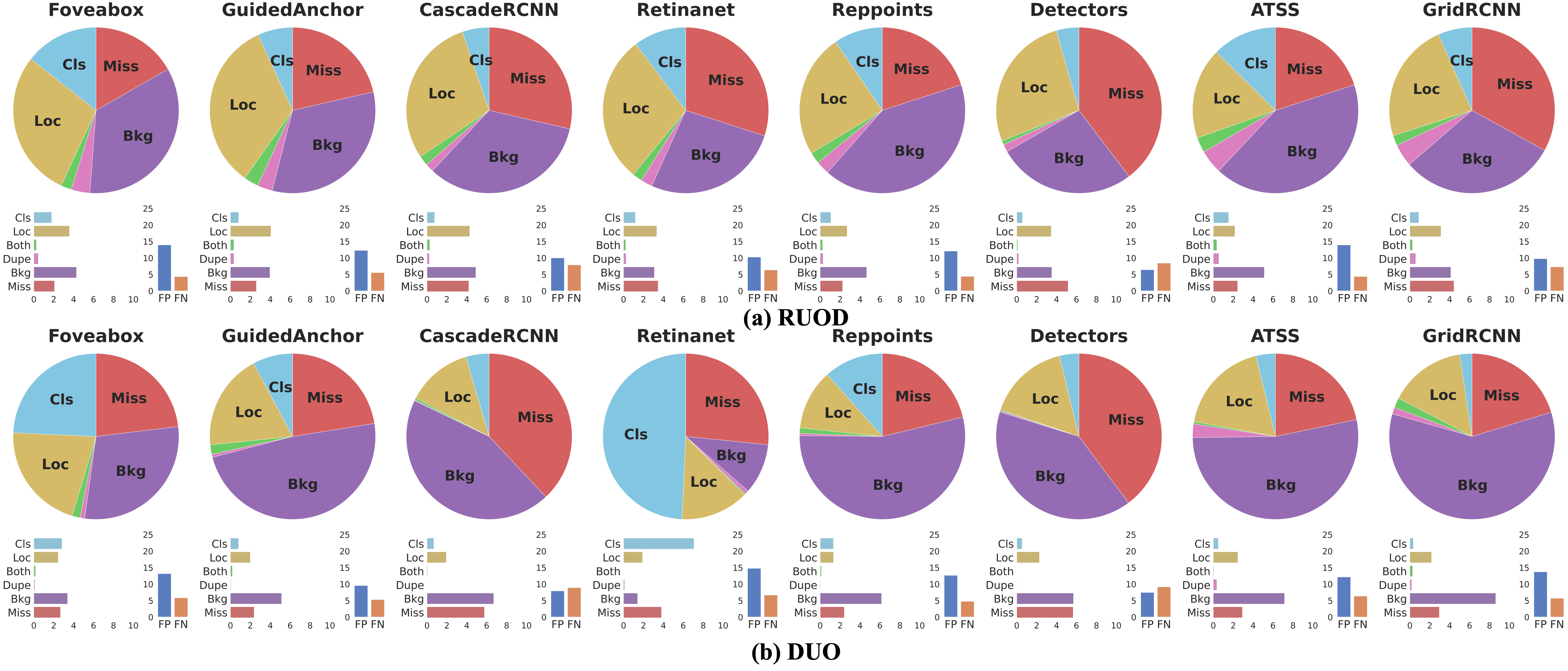}
\caption{The error types of representative object detection frameworks on (a) RUOD and (b) DUO. The pie chart represents the relative contribution of each error type, while the bar plots display their absolute contribution.}
\label{fig:object_error}
\end{figure*}

Diagnosis well-examines the impact of object characteristics, such as size and aspect ratio, on detection performance. Diagnosis categorises objects into five size groups based on their percentile size within each object category: extra-small (XS: bottom 10\%), small (S: next 20\%), medium (M: next 40\%), large (L: next 20\%), and extra-large (XL: top 10\%), as illustrated in Fig. \ref{fig:object_size_RUOD}. Similarly, aspect ratio, defined as the object's width divided by its height, is categorised into five groups based on the percentile rankings: extra-tall (XT: bottom 10\%), tall (T: next 20\%), medium (M: next 40\%), wide (W: next 20\%), and extra-wide (XW: top 10\%), as illustrated in Fig. \ref{fig:object_ar_RUOD}. TIDE offers a more precise and comprehensive error analysis than Diagnosis. It includes six error types: classification error, localisation error, both classification and localisation error, duplicate detection error, background error, and missed GT error. Fig. \ref{fig:object_error} illustrates the distributions of error types as defined by TIDE. The pie chart represents the relative contribution of each error type, while the bar plots display their absolute contribution. To support research in underwater object detection, we publish ready-to-use source code for Diagnosis and Tide, specifically designed for the RUOD and DUO datasets.

\section{Evaluations of mainstream detection frameworks on the RUOD and DUO benchmarks}
\label{sec:experiments}

Although various datasets have been used to evaluate various UOD algorithms, well-recognised and reliable benchmarks are still urgently needed to support UOD development for two reasons: (1) researchers often use the same dataset but apply different splits, particularly with the URPC series datasets, which lack test set annotations, and (2) many early-stage datasets, like URPC2017 and URPC2018, contain considerable annotation errors. The results reported on test sets with faulty annotations are neither fair nor reliable. Therefore, we recommend two high-quality, large-scale datasets, RUOD and DUO, with consistent data split and accurate annotations as unified benchmarks for comparing UOD algorithms.

Many specialised underwater detectors have not released their source codes or their environments are difficulty to reproduce. While many generic detectors have been tested on the RUOD and DUO datasets, the trained models and detection results are not publicly accessible. As a result, researchers must repeat experiments, causing duplicated, time-consuming efforts. To address these issues, we introduce UODReview, a platform specially designed for comparing various underwater object detection approaches. It provides source codes, trained models, detection results, and detection analysis tools for RUOD and DUO, facilitating faster comparisons and advancing UOD development. Both mainstream generic detectors and SOAT underwater detectors are chosen for comparison. For the generic detectors, we utilise source codes from MMDetection to retrain and evaluate several leading detectors, including RetinaNet \cite{lin2017focal}, FoveaBox \cite{kong2020foveabox}, ATSS \cite{zhang2020bridging}, GridRCNN \cite{lu2019grid}, DetectoRS \cite{qiao2021detectors}, RepPoints \cite{yang2019reppoints}, CascadeRCNN \cite{cai2019cascade}, and GuidedAnchor \cite{wang2019region}. For the specialised underwater detectors, we present results reported in previous studies directly, as most of these detectors have not published their source code or cannot be successfully reproduced. On the DUO and RUOD datasets, we consider several top-performing methods, including DJLNet \cite{wang2024dual}, ERLNet \cite{dai2023edge}, GCCNet \cite{dai2024gated}, RoIMix \cite{lin2020roimix}, BoostRCNN \cite{song2023boosting}, RFTM \cite{fu2023learning} and RoIAttn \cite{liang2022excavating}. Their results are sourced directly from the study by Wang et al. \cite{wang2024dual}.

\subsection{Comparisons between Underwater and Generic Detectors}

Table \ref{tab:quantitative} presents the quantitative results, while Fig. \ref{fig:object_imbalance_RUOD} displays the precision-recall curves for underwater detectors and generic detectors. From these, we observe that many underwater detectors, such as DJNet, GCCNet, and ERLNet, performs much better than generic detectors. Many generic detectors, especially FoveaBox and GuidedAnchor on RUOD, and RetinaNet and FoveaBox on DUO, performs much worse. This is primarily due to several key factors related to the unique challenges and characteristics of the underwater environment. For instance, DJLNet excels at simultaneously learning both appearance and edge features, enhancing robustness in complex underwater scenes. It incorporates an image decolorisation module to eliminate color distortions caused by underwater light absorption and scattering. Additionally, its edge enhancement branch sharpens shape and texture details, improving the recognition of object boundary features. Similarly, ERLNet leverages an edge‐guided attention module to refine edge information for better detection performance. By sharpening object boundaries, it enhances the detection of objects with fuzzy boundaries, a common challenge in underwater images due to blurring and low visibility. GCCNet tackles the challenges of poor visibility and low contrast in underwater environments through three key components: first, a UIE method that enhances object visibility in low-contrast areas; second, a cross-domain feature interaction module that extracts complementary information between raw and enhanced images; and third, a gated feature fusion module that adaptively manages the fusion ratio of cross-domain information.

Generic detectors often exhibit poor or unstable performance across different underwater datasets. For instance, GuidedAnchor perform well on the DUO dataset but far lags behind other methods on the RUOD dataset. Similarly, CascadeRCNN shows varying performance on two datasets. This discrepancy is largely due to domain-specific variations in the datasets: RUOD features diverse object sizes, color distortions, and lighting conditions, whereas the DUO dataset has a more consistent style. Cascade R-CNN enhances object detection through a multi-stage framework where each stage progressively refines the bounding box predictions. This approach enhances detection quality, particularly for challenging cases such as small or occluded objects, however, it can also overfit on easier-to-detect objects, especially when more refinement stages are used. GuidedAnchor presents a new anchor generation method that uses feature-guided anchors to improve object localisation. However, it can be less effective for objects with extreme aspect ratios or variable shapes, leading to challenges in generalising across diverse datasets. Hence, the design of underwater detectors should consider both the unique characteristics of the underwater environment and the variations present in different datasets.

\subsection{Discussions on the Impact of Object Characteristics}

\begin{figure*}[h]
\centering
\includegraphics[width=18cm]{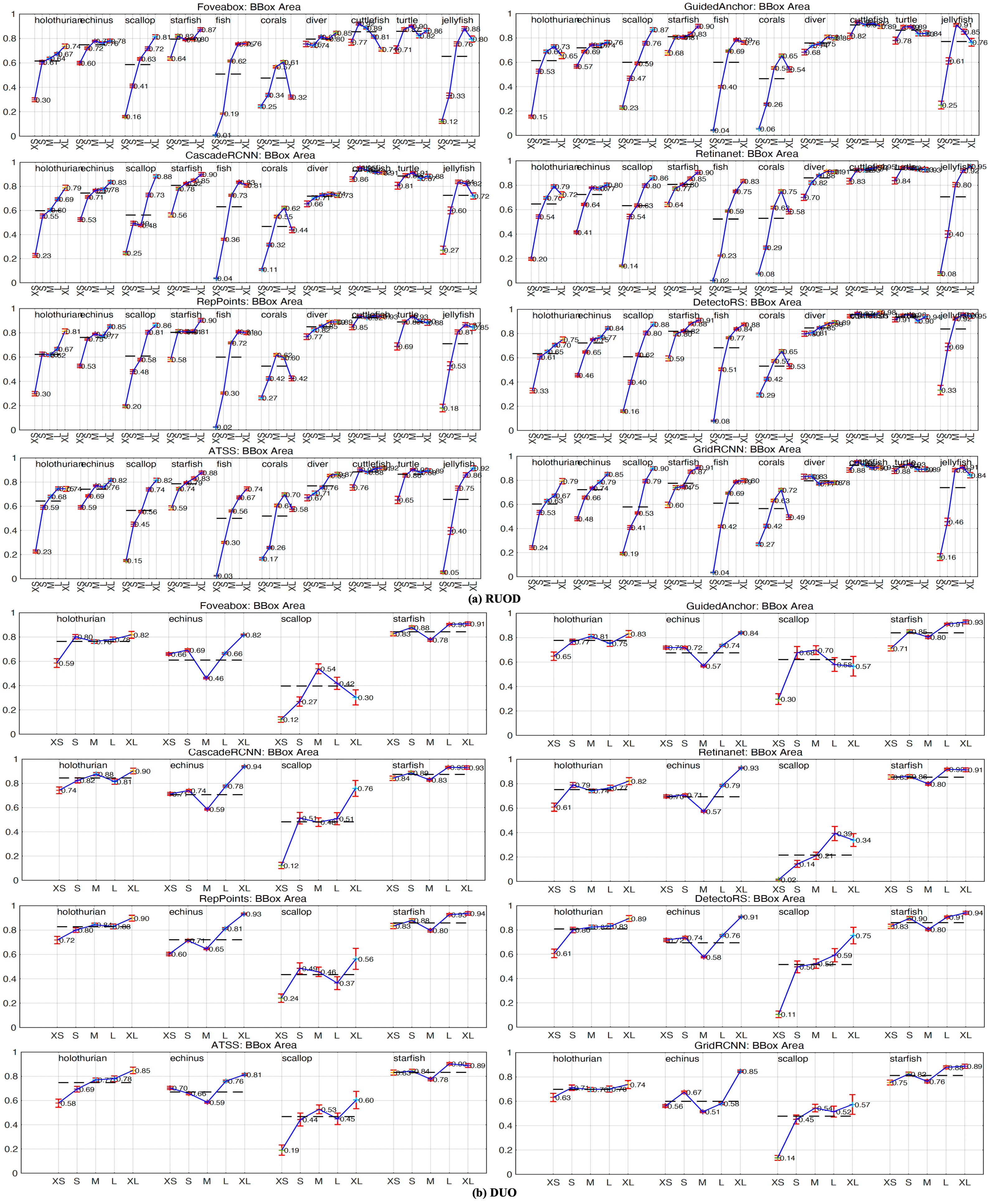}
\caption{Performance of representative detection frameworks for objects of varying sizes in the RUOD and DUO datasets: XS (bottom 10\%)=extra-small, S (next 20\%)=small, M (next 40\%)=medium, L (next 20\%)=large, XL (top 10\%)=extra-large.}
\label{fig:object_size_RUOD}
\end{figure*}

Fig. \ref{fig:object_size_RUOD} presents the performance of representative detection frameworks for objects of different sizes on the RUOD and DUO datasets. In both datasets, object characteristics, such as size and aspect ratio, significantly impact detection performance. Most detection frameworks exhibit very low precision for the extra-small object category, as these objects occupy only a few pixels and contain much less visual information, making it challenging for detectors to distinguish them from backgrounds and noise in underwater images. Moreover, current detectors strive to maintain scale invariance, with detection accuracy varying largely for objects of different sizes. Detecting a wide range of object scales is challenging because models must remain sensitive to small objects without sacrificing performance on larger ones.

\begin{figure*}[h]
\centering
\includegraphics[width=18cm]{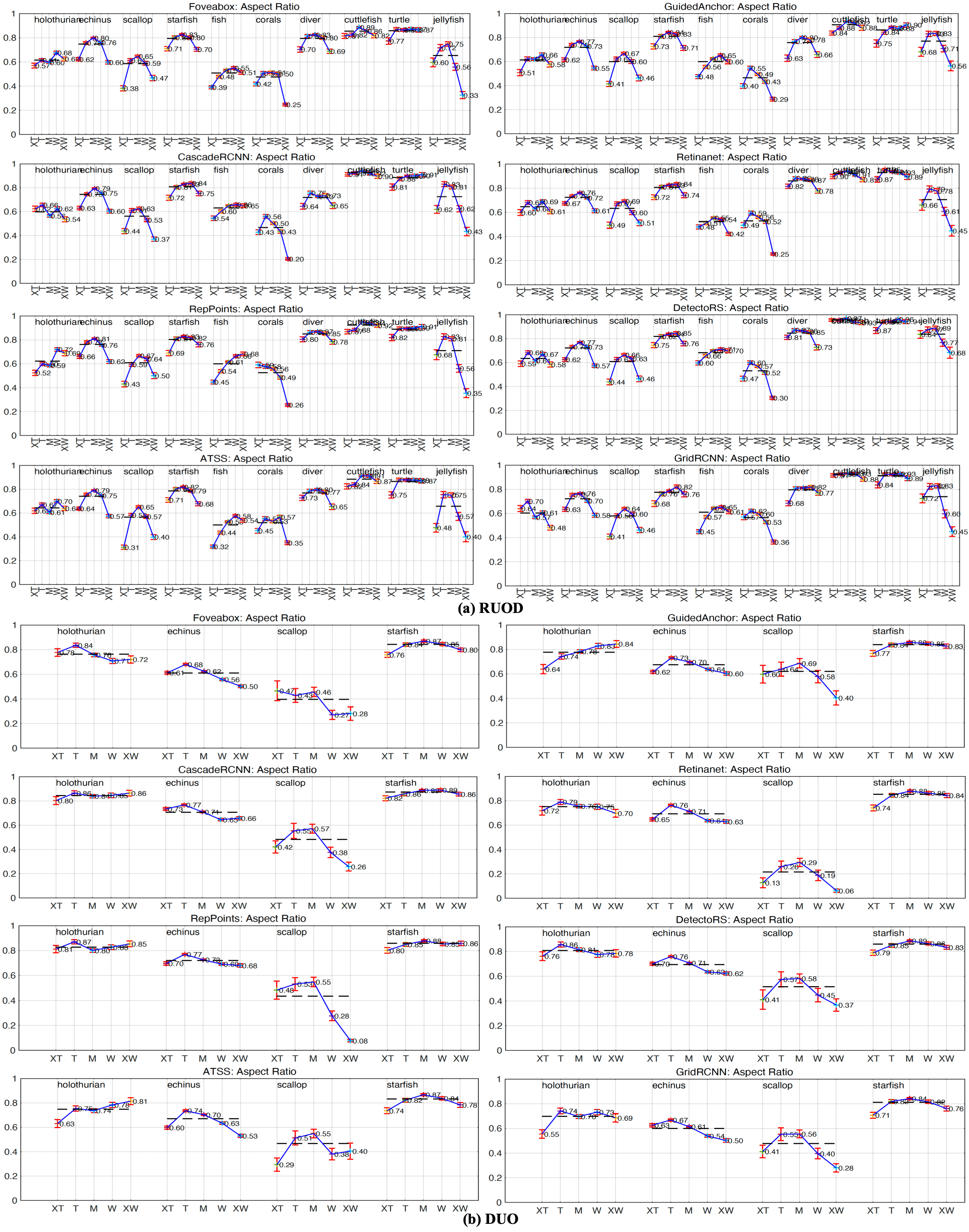}
\caption{Performance of representative detection frameworks for objects of varying aspect ratios in the RUOD and DUO datasets: XT (bottom 10\%)=extra-tall, T (next 20\%)=tall, M (next 40\%)=medium, W (next 20\%)=wide, XW (top 10\%)=extra-wide.}
\label{fig:object_ar_RUOD}
\end{figure*}

Fig. \ref{fig:object_ar_RUOD} presents the performance of representative detection frameworks for objects with different aspect ratios on the RUOD and DUO datasets. Most detectors prefer objects with a 1:1 aspect ratio (i.e., squared objects). This can be attributed due to several aspects of detection network architecture: (1) Square Kernels-Most detectors utilise square convolutional filters to process input images. These filters are inherently more effective at capturing patterns in objects closer to a 1:1 aspect ratio, as the spatial information is processed evenly in both dimensions (height and width). (2) Anchors Boxes-Many detectors rely on anchor boxes to predict object locations and categories. These anchor boxes are predefined with various aspect ratios, but 1:1 aspect ratios are commonly emphasized. Objects with extreme aspect ratios (either extremely tall or extremely wide) may not align well with these square filters, and the predefined anchor boxes may not match their shapes, leading to less accurate predictions.

\subsection{Discussions on the Impact of Error Types}

Fig. \ref{fig:object_error} presents the error types of representative object detection frameworks on these two datasets. On the RUOD dataset, most detectors, including Foveabox, CascadeRCNN, Reppoints, and ATSS, tend to make more background and localisation errors compared to other error types. This is due 

\clearpage
\noindent to severe color distortions that obscure the distinctions between objects and backgrounds. DetectoRS and GridRCNN exhibit a higher number of missed ground truth errors, indicating that many objects are miss detected. GridRCNN utilises a grid-guided localisation approach to refine bounding boxes, which, although precise, can struggle with objects that have low contrast and irregular shapes (e.g., seacucumbers). The grid representation can sometimes fail to capture objects, especially when they are small or blurred. DetectoRS employs advanced feature extraction techniques, such as Recursive Feature Pyramid (RFP) and Switchable Atrous Convolution (SAC), which significantly enhance feature representation but also increase model complexity. This complexity can occasionally cause important features to be underrepresented or masked by noise during the recursive processing, leading to missed detection.

On the DUO dataset, the background errors are more pronounced for most of the detectors, mainly due to blurring effects that cause seacucumbers to resemble sand and scallops to appear like stones, as illustrated in Fig. \ref{fig:example_im}. This significantly complicates accurate detection. It is important to note that Retinanet produces more classification errors. This may be attributed to its use of focal loss to address class imbalance. Focal loss emphasises hard-to-classify examples, which can cause the model to overlook easier examples. If the model becomes overly focused these challenging cases, it may misclassify simply instances, resulting in an increase in overall errors.

\section{Future insights and vision}
\label{sec:conclusions}

This survey presents a systematic review of AI-based UOD techniques. Our analysis reveals that sonar images are better choices than RGB images for UOD in extremely murky waters due to their superior perceptual range. Furthermore, developing improved networks architectures, loss functions, and learning strategies are promising directions to enhance UOD performance. We explore the challenges in UOD, noting that some issues, such as noisy labels, have not been thoroughly investigated, and that advanced techniques from GOD could help address these challenges. Additionally, we provide two valuable detection analysis tools to help identify the strengths and weaknesses of each underwater detector. Finally, we recommend two high-quality, large-scale benchmarks for fair comparison of UOD algorithms. Several mainstream deep detectors are evaluated on these benchmarks, and the source codes, trained models, utilised datasets, and detection results are available online, enabling researchers to compare their detectors against existing methods.

\ifCLASSOPTIONcaptionsoff
  \newpage
\fi
\bibliographystyle{IEEEtran}
\bibliography{mybibfile}

\begin{thebibliography}{100}
\providecommand{\url}[1]{#1}
\csname url@samestyle\endcsname
\providecommand{\newblock}{\relax}
\providecommand{\bibinfo}[2]{#2}
\providecommand{\BIBentrySTDinterwordspacing}{\spaceskip=0pt\relax}
\providecommand{\BIBentryALTinterwordstretchfactor}{4}
\providecommand{\BIBentryALTinterwordspacing}{\spaceskip=\fontdimen2\font plus
\BIBentryALTinterwordstretchfactor\fontdimen3\font minus
  \fontdimen4\font\relax}
\providecommand{\BIBforeignlanguage}[2]{{%
\expandafter\ifx\csname l@#1\endcsname\relax
\typeout{** WARNING: IEEEtran.bst: No hyphenation pattern has been}%
\typeout{** loaded for the language `#1'. Using the pattern for}%
\typeout{** the default language instead.}%
\else
\language=\csname l@#1\endcsname
\fi
#2}}
\providecommand{\BIBdecl}{\relax}
\BIBdecl

\bibitem{chiang2011underwater}
J.~Y. Chiang and Y.-C. Chen, ``Underwater image enhancement by wavelength
  compensation and dehazing,'' \emph{IEEE Transactions on Image Processing},
  vol.~21, no.~4, pp. 1756--1769, 2011.

\bibitem{fu2023rethinking}
C.~Fu, R.~Liu, X.~Fan, P.~Chen, H.~Fu, W.~Yuan, M.~Zhu, and Z.~Luo,
  ``Rethinking general underwater object detection: Datasets, challenges, and
  solutions,'' \emph{Neurocomputing}, vol. 517, pp. 243--256, 2023.

\bibitem{yang2019reppoints}
Z.~Yang, S.~Liu, H.~Hu, L.~Wang, and S.~Lin, ``Reppoints: Point set
  representation for object detection,'' in \emph{Proceedings of the IEEE/CVF
  International Conference on Computer Vision}, 2019, pp. 9657--9666.

\bibitem{liu2021dataset}
C.~Liu, H.~Li, S.~Wang, M.~Zhu, D.~Wang, X.~Fan, and Z.~Wang, ``A dataset and
  benchmark of underwater object detection for robot picking,'' in \emph{2021
  IEEE International Conference on Multimedia \& Expo Workshops (ICMEW)}.\hskip
  1em plus 0.5em minus 0.4em\relax IEEE, 2021, pp. 1--6.

\bibitem{akkaynak2018revised}
D.~Akkaynak and T.~Treibitz, ``A revised underwater image formation model,'' in
  \emph{Proceedings of the IEEE Conference on Computer Vision and Pattern
  Recognition}, 2018, pp. 6723--6732.

\bibitem{xu2023systematic}
S.~Xu, M.~Zhang, W.~Song, H.~Mei, Q.~He, and A.~Liotta, ``A systematic review
  and analysis of deep learning-based underwater object detection,''
  \emph{Neurocomputing}, 2023.

\bibitem{ren2016faster}
S.~Ren, K.~He, R.~Girshick, and J.~Sun, ``Faster r-cnn: Towards real-time
  object detection with region proposal networks,'' \emph{IEEE Transactions on
  Pattern Analysis and Machine Intelligence}, vol.~39, no.~6, pp. 1137--1149,
  2016.

\bibitem{liu2016ssd}
W.~Liu, D.~Anguelov, D.~Erhan, C.~Szegedy, S.~Reed, C.-Y. Fu, and A.~C. Berg,
  ``Ssd: Single shot multibox detector,'' in \emph{European Conference on
  Computer Vision}.\hskip 1em plus 0.5em minus 0.4em\relax Springer, 2016, pp.
  21--37.

\bibitem{redmon2016you}
J.~Redmon, S.~Divvala, R.~Girshick, and A.~Farhadi, ``You only look once:
  Unified, real-time object detection,'' in \emph{Proceedings of the IEEE
  Conference on Computer Vision and Pattern Recognition}, 2016, pp. 779--788.

\bibitem{hoiem2012diagnosing}
D.~Hoiem, Y.~Chodpathumwan, and Q.~Dai, ``Diagnosing error in object
  detectors,'' in \emph{European Conference on Computer Vision}.\hskip 1em plus
  0.5em minus 0.4em\relax Springer, 2012, pp. 340--353.

\bibitem{bolya2020tide}
D.~Bolya, S.~Foley, J.~Hays, and J.~Hoffman, ``Tide: A general toolbox for
  identifying object detection errors,'' in \emph{Computer Vision--ECCV 2020:
  16th European Conference, Glasgow, UK, August 23--28, 2020, Proceedings, Part
  III 16}.\hskip 1em plus 0.5em minus 0.4em\relax Springer, 2020, pp. 558--573.

\bibitem{moniruzzaman2017deep}
M.~Moniruzzaman, S.~M.~S. Islam, M.~Bennamoun, and P.~Lavery, ``Deep learning
  on underwater marine object detection: A survey,'' in \emph{Advanced Concepts
  for Intelligent Vision Systems: 18th International Conference, ACIVS 2017,
  Antwerp, Belgium, September 18-21, 2017, Proceedings 18}.\hskip 1em plus
  0.5em minus 0.4em\relax Springer, 2017, pp. 150--160.

\bibitem{gomes2020robust}
D.~Gomes, A.~S. Saif, and D.~Nandi, ``Robust underwater object detection with
  autonomous underwater vehicle: A comprehensive study,'' in \emph{Proceedings
  of the International Conference on Computing Advancements}, 2020, pp. 1--10.

\bibitem{wang2022review}
N.~Wang, Y.~Wang, and M.~J. Er, ``Review on deep learning techniques for marine
  object recognition: Architectures and algorithms,'' \emph{Control Engineering
  Practice}, vol. 118, p. 104458, 2022.

\bibitem{fayaz2022underwater}
S.~Fayaz, S.~A. Parah, and G.~Qureshi, ``Underwater object detection:
  architectures and algorithms--a comprehensive review,'' \emph{Multimedia
  Tools and Applications}, vol.~81, no.~15, pp. 20\,871--20\,916, 2022.

\bibitem{jian2024underwater}
M.~Jian, N.~Yang, C.~Tao, H.~Zhi, and H.~Luo, ``Underwater object detection and
  datasets: a survey,'' \emph{Intelligent Marine Technology and Systems},
  vol.~2, no.~1, p.~9, 2024.

\bibitem{viola2001rapid}
P.~Viola and M.~Jones, ``Rapid object detection using a boosted cascade of
  simple features,'' in \emph{Proceedings of the 2001 IEEE Computer Society
  Conference on Computer Vision and Pattern Recognition. CVPR 2001},
  vol.~1.\hskip 1em plus 0.5em minus 0.4em\relax Ieee, 2001, pp. I--I.

\bibitem{dalal2005histograms}
N.~Dalal and B.~Triggs, ``Histograms of oriented gradients for human
  detection,'' in \emph{2005 IEEE Computer Society Conference on Computer
  Vision and Pattern Recognition (CVPR'05)}, vol.~1.\hskip 1em plus 0.5em minus
  0.4em\relax Ieee, 2005, pp. 886--893.

\bibitem{felzenszwalb2008discriminatively}
P.~Felzenszwalb, D.~McAllester, and D.~Ramanan, ``A discriminatively trained,
  multiscale, deformable part model,'' in \emph{2008 IEEE Conference on
  Computer Vision and Pattern Recognition}.\hskip 1em plus 0.5em minus
  0.4em\relax Ieee, 2008, pp. 1--8.

\bibitem{girshick2014rich}
R.~Girshick, J.~Donahue, T.~Darrell, and J.~Malik, ``Rich feature hierarchies
  for accurate object detection and semantic segmentation,'' in
  \emph{Proceedings of the IEEE Conference on Computer Vision and Pattern
  Recognition}, 2014, pp. 580--587.

\bibitem{dai2016r}
J.~Dai, Y.~Li, K.~He, and J.~Sun, ``R-fcn: Object detection via region-based
  fully convolutional networks,'' \emph{Advances in Neural Information
  Processing Systems}, vol.~29, 2016.

\bibitem{lin2017feature}
T.-Y. Lin, P.~Doll{\'a}r, R.~Girshick, K.~He, B.~Hariharan, and S.~Belongie,
  ``Feature pyramid networks for object detection,'' in \emph{Proceedings of
  the IEEE Conference on Computer Vision and Pattern Recognition}, 2017, pp.
  2117--2125.

\bibitem{lin2017focal}
T.-Y. Lin, P.~Goyal, R.~Girshick, K.~He, and P.~Doll{\'a}r, ``Focal loss for
  dense object detection,'' in \emph{Proceedings of the IEEE International
  Conference on Computer Vision}, 2017, pp. 2980--2988.

\bibitem{law2018cornernet}
H.~Law and J.~Deng, ``Cornernet: Detecting objects as paired keypoints,'' in
  \emph{Proceedings of the European Conference on Computer Vision (ECCV)},
  2018, pp. 734--750.

\bibitem{duan2019centernet}
K.~Duan, S.~Bai, L.~Xie, H.~Qi, Q.~Huang, and Q.~Tian, ``Centernet: Keypoint
  triplets for object detection,'' in \emph{Proceedings of the IEEE/CVF
  International Conference on Computer Vision}, 2019, pp. 6569--6578.

\bibitem{carion2020end}
N.~Carion, F.~Massa, G.~Synnaeve, N.~Usunier, A.~Kirillov, and S.~Zagoruyko,
  ``End-to-end object detection with transformers,'' in \emph{Computer
  Vision--ECCV 2020: 16th European Conference, Glasgow, UK, August 23--28,
  2020, Proceedings, Part I 16}.\hskip 1em plus 0.5em minus 0.4em\relax
  Springer, 2020, pp. 213--229.

\bibitem{zhu2020deformable}
X.~Zhu, W.~Su, L.~Lu, B.~Li, X.~Wang, and J.~Dai, ``Deformable detr: Deformable
  transformers for end-to-end object detection,'' \emph{arXiv preprint
  arXiv:2010.04159}, 2020.

\bibitem{kim2014artificial}
D.~Kim, D.~Lee, H.~Myung, and H.-T. Choi, ``Artificial landmark-based
  underwater localization for auvs using weighted template matching,''
  \emph{Intelligent Service Robotics}, vol.~7, no.~3, pp. 175--184, 2014.

\bibitem{strachan1993recognition}
N.~Strachan, ``Recognition of fish species by colour and shape,'' \emph{Image
  and Vision Computing}, vol.~11, no.~1, pp. 2--10, 1993.

\bibitem{marcos2005classification}
M.~S. A.~C. Marcos, M.~N. Soriano, and C.~A. Saloma, ``Classification of coral
  reef images from underwater video using neural networks,'' \emph{Optics
  Express}, vol.~13, no.~22, pp. 8766--8771, 2005.

\bibitem{barat2006robust}
C.~Barat and M.-J. Rendas, ``A robust visual attention system for detecting
  manufactured objects in underwater video,'' in \emph{OCEANS 2006}.\hskip 1em
  plus 0.5em minus 0.4em\relax IEEE, 2006, pp. 1--6.

\bibitem{chew2007automatic}
A.~L. Chew, P.~B. Tong, and C.~S. Chia, ``Automatic detection and
  classification of man-made targets in side scan sonar images,'' in \emph{2007
  Symposium on Underwater Technology and Workshop on Scientific Use of
  Submarine Cables and Related Technologies}.\hskip 1em plus 0.5em minus
  0.4em\relax IEEE, 2007, pp. 126--132.

\bibitem{spampinato2008detecting}
C.~Spampinato, Y.-H. Chen-Burger, G.~Nadarajan, and R.~B. Fisher, ``Detecting,
  tracking and counting fish in low quality unconstrained underwater videos.''
  \emph{VISAPP (2)}, vol. 2008, no. 514-519, p.~1, 2008.

\bibitem{beijbom2012automated}
O.~Beijbom, P.~J. Edmunds, D.~I. Kline, B.~G. Mitchell, and D.~Kriegman,
  ``Automated annotation of coral reef survey images,'' in \emph{2012 IEEE
  Conference on Computer Vision and Pattern Recognition}.\hskip 1em plus 0.5em
  minus 0.4em\relax IEEE, 2012, pp. 1170--1177.

\bibitem{spampinato2012covariance}
C.~Spampinato, S.~Palazzo, D.~Giordano, I.~Kavasidis, F.-P. Lin, and Y.-T. Lin,
  ``Covariance based fish tracking in real-life underwater environment.'' in
  \emph{VISAPP (2)}, 2012, pp. 409--414.

\bibitem{galceran2012real}
E.~Galceran, V.~Djapic, M.~Carreras, and D.~P. Williams, ``A real-time
  underwater object detection algorithm for multi-beam forward looking sonar,''
  \emph{IFAC Proceedings Volumes}, vol.~45, no.~5, pp. 306--311, 2012.

\bibitem{li2013underwater}
M.~Li, H.~Ji, X.~Wang, L.~Weng, and Z.~Gong, ``Underwater object detection and
  tracking based on multi-beam sonar image processing,'' in \emph{2013 IEEE
  International Conference on Robotics and Biomimetics (ROBIO)}.\hskip 1em plus
  0.5em minus 0.4em\relax IEEE, 2013, pp. 1071--1076.

\bibitem{lee2004contour}
D.-J. Lee, R.~B. Schoenberger, D.~Shiozawa, X.~Xu, and P.~Zhan, ``Contour
  matching for a fish recognition and migration-monitoring system,'' in
  \emph{Two-and Three-Dimensional Vision Systems for Inspection, Control, and
  Metrology II}, vol. 5606.\hskip 1em plus 0.5em minus 0.4em\relax
  International Society for Optics and Photonics, 2004, pp. 37--48.

\bibitem{ravanbakhsh2015automated}
M.~Ravanbakhsh, M.~R. Shortis, F.~Shafait, A.~Mian, E.~S. Harvey, and J.~W.
  Seager, ``Automated fish detection in underwater images using shape-based
  level sets,'' \emph{The Photogrammetric Record}, vol.~30, no. 149, pp.
  46--62, 2015.

\bibitem{cho2015acoustic}
H.~Cho, J.~Gu, H.~Joe, A.~Asada, and S.-C. Yu, ``Acoustic beam profile-based
  rapid underwater object detection for an imaging sonar,'' \emph{Journal of
  Marine Science and Technology}, vol.~20, pp. 180--197, 2015.

\bibitem{hou2016underwater}
G.-J. Hou, X.~Luan, D.-L. Song, and X.-Y. Ma, ``Underwater man-made object
  recognition on the basis of color and shape features,'' \emph{Journal of
  Coastal Research}, vol.~32, no.~5, pp. 1135--1141, 2016.

\bibitem{chuang2016feature}
M.-C. Chuang, J.-N. Hwang, and K.~Williams, ``A feature learning and object
  recognition framework for underwater fish images,'' \emph{IEEE Transactions
  on Image Processing}, vol.~25, no.~4, pp. 1862--1872, 2016.

\bibitem{villon2016coral}
S.~Villon, M.~Chaumont, G.~Subsol, S.~Vill{\'e}ger, T.~Claverie, and
  D.~Mouillot, ``Coral reef fish detection and recognition in underwater videos
  by supervised machine learning: Comparison between deep learning and hog+ svm
  methods,'' in \emph{International Conference on Advanced Concepts for
  Intelligent Vision Systems}.\hskip 1em plus 0.5em minus 0.4em\relax Springer,
  2016, pp. 160--171.

\bibitem{liu2016combining}
H.~Liu, J.~Dai, R.~Wang, H.~Zheng, and B.~Zheng, ``Combining background
  subtraction and three-frame difference to detect moving object from
  underwater video,'' in \emph{OCEANS 2016-Shanghai}.\hskip 1em plus 0.5em
  minus 0.4em\relax IEEE, 2016, pp. 1--5.

\bibitem{chen2017monocular}
Z.~Chen, Z.~Zhang, F.~Dai, Y.~Bu, and H.~Wang, ``Monocular vision-based
  underwater object detection,'' \emph{Sensors}, vol.~17, no.~8, p. 1784, 2017.

\bibitem{kim2017imaging}
B.~Kim and S.-C. Yu, ``Imaging sonar based real-time underwater object
  detection utilizing adaboost method,'' in \emph{2017 IEEE Underwater
  Technology (UT)}.\hskip 1em plus 0.5em minus 0.4em\relax IEEE, 2017, pp.
  1--5.

\bibitem{vasamsetti2018automatic}
S.~Vasamsetti, S.~Setia, N.~Mittal, H.~K. Sardana, and G.~Babbar, ``Automatic
  underwater moving object detection using multi-feature integration framework
  in complex backgrounds,'' \emph{IET Computer Vision}, vol.~12, no.~6, pp.
  770--778, 2018.

\bibitem{yu2021real}
Y.~Yu, J.~Zhao, Q.~Gong, C.~Huang, G.~Zheng, and J.~Ma, ``Real-time underwater
  maritime object detection in side-scan sonar images based on
  transformer-yolov5,'' \emph{Remote Sensing}, vol.~13, no.~18, p. 3555, 2021.

\bibitem{barngrover2015brain}
C.~Barngrover, A.~Althoff, P.~DeGuzman, and R.~Kastner, ``A brain--computer
  interface (bci) for the detection of mine-like objects in sidescan sonar
  imagery,'' \emph{IEEE Journal of Oceanic Engineering}, vol.~41, no.~1, pp.
  123--138, 2015.

\bibitem{zhang2022target}
H.~Zhang, M.~Tian, G.~Shao, J.~Cheng, and J.~Liu, ``Target detection of
  forward-looking sonar image based on improved yolov5,'' \emph{IEEE Access},
  vol.~10, pp. 18\,023--18\,034, 2022.

\bibitem{zhou2022automatic}
T.~Zhou, J.~Si, L.~Wang, C.~Xu, and X.~Yu, ``Automatic detection of underwater
  small targets using forward-looking sonar images,'' \emph{IEEE Transactions
  on Geoscience and Remote Sensing}, vol.~60, pp. 1--12, 2022.

\bibitem{hayes1992broad}
M.~P. Hayes and P.~T. Gough, ``Broad-band synthetic aperture sonar,''
  \emph{IEEE Journal of Oceanic Engineering}, vol.~17, no.~1, pp. 80--94, 1992.

\bibitem{blanc2014fish}
K.~Blanc, D.~Lingrand, and F.~Precioso, ``Fish species recognition from video
  using svm classifier,'' in \emph{Proceedings of the 3rd ACM International
  Workshop on Multimedia Analysis for Ecological Data}, 2014, pp. 1--6.

\bibitem{bay2006surf}
H.~Bay, T.~Tuytelaars, and L.~Van~Gool, ``Surf: Speeded up robust features,''
  \emph{Lecture Notes in Computer Science}, vol. 3951, pp. 404--417, 2006.

\bibitem{li2015fast}
X.~Li, M.~Shang, H.~Qin, and L.~Chen, ``Fast accurate fish detection and
  recognition of underwater images with fast r-cnn,'' in \emph{OCEANS
  2015-MTS/IEEE Washington}.\hskip 1em plus 0.5em minus 0.4em\relax IEEE, 2015,
  pp. 1--5.

\bibitem{shkurti2017underwater}
F.~Shkurti, W.-D. Chang, P.~Henderson, M.~J. Islam, J.~C.~G. Higuera, J.~Li,
  T.~Manderson, A.~Xu, G.~Dudek, and J.~Sattar, ``Underwater multi-robot
  convoying using visual tracking by detection,'' in \emph{2017 IEEE/RSJ
  International Conference on Intelligent Robots and Systems (IROS)}.\hskip 1em
  plus 0.5em minus 0.4em\relax IEEE, 2017, pp. 4189--4196.

\bibitem{li2017deep}
X.~Li, Y.~Tang, and T.~Gao, ``Deep but lightweight neural networks for fish
  detection,'' in \emph{OCEANS 2017-Aberdeen}.\hskip 1em plus 0.5em minus
  0.4em\relax IEEE, 2017, pp. 1--5.

\bibitem{ji2018design}
L.~Ji-Yong, Z.~Hao, H.~Hai, Y.~Xu, W.~Zhaoliang, and W.~Lei, ``Design and
  vision based autonomous capture of sea organism with absorptive type remotely
  operated vehicle,'' \emph{IEEE Access}, vol.~6, pp. 73\,871--73\,884, 2018.

\bibitem{zhang2018single}
L.~Zhang, X.~Yang, Z.~Liu, L.~Qi, H.~Zhou, and C.~Chiu, ``Single shot feature
  aggregation network for underwater object detection,'' in \emph{2018 24th
  International Conference on Pattern Recognition (ICPR)}.\hskip 1em plus 0.5em
  minus 0.4em\relax IEEE, 2018, pp. 1906--1911.

\bibitem{lee2018deep}
S.~Lee, B.~Park, and A.~Kim, ``Deep learning from shallow dives: Sonar image
  generation and training for underwater object detection,'' \emph{arXiv
  preprint arXiv:1810.07990}, 2018.

\bibitem{pedersen2019detection}
M.~Pedersen, J.~Bruslund~Haurum, R.~Gade, and T.~B. Moeslund, ``Detection of
  marine animals in a new underwater dataset with varying visibility,'' in
  \emph{Proceedings of the IEEE/CVF Conference on Computer Vision and Pattern
  Recognition Workshops}, 2019, pp. 18--26.

\bibitem{long2020underwater}
L.~Chen, Z.~Liu, L.~Tong, Z.~Jiang, S.~Wang, J.~Dong, and H.~Zhou, ``Underwater
  object detection using invert multi-class adaboost with deep learning,'' in
  \emph{2020 International Joint Conference on Neural Networks, IJCNN
  2020}.\hskip 1em plus 0.5em minus 0.4em\relax Institute of Electrical and
  Electronics Engineers (IEEE), 2020.

\bibitem{zhang2020research}
J.~Zhang, L.~Zhu, L.~Xu, and Q.~Xie, ``Research on the correlation between
  image enhancement and underwater object detection,'' in \emph{2020 Chinese
  Automation Congress (CAC)}.\hskip 1em plus 0.5em minus 0.4em\relax IEEE,
  2020, pp. 5928--5933.

\bibitem{chen2020underwater}
W.~Chen and B.~Fan, ``Underwater object detection with mixed attention
  mechanism and multi-enhancement strategy,'' in \emph{2020 Chinese Automation
  Congress (CAC)}.\hskip 1em plus 0.5em minus 0.4em\relax IEEE, 2020, pp.
  2821--2826.

\bibitem{wang2020yolo}
L.~Wang, X.~Ye, H.~Xing, Z.~Wang, and P.~Li, ``Yolo nano underwater: A fast and
  compact object detector for embedded device,'' in \emph{Global Oceans 2020:
  Singapore--US Gulf Coast}.\hskip 1em plus 0.5em minus 0.4em\relax IEEE, 2020,
  pp. 1--4.

\bibitem{liu2020towards}
H.~Liu, P.~Song, and R.~Ding, ``Towards domain generalization in underwater
  object detection,'' in \emph{2020 IEEE International Conference on Image
  Processing (ICIP)}.\hskip 1em plus 0.5em minus 0.4em\relax IEEE, 2020, pp.
  1971--1975.

\bibitem{fan2020dual}
B.~Fan, W.~Chen, Y.~Cong, and J.~Tian, ``Dual refinement underwater object
  detection network,'' in \emph{Computer Vision--ECCV 2020: 16th European
  Conference, Glasgow, UK, August 23--28, 2020, Proceedings, Part XX 16}.\hskip
  1em plus 0.5em minus 0.4em\relax Springer, 2020, pp. 275--291.

\bibitem{zhang2020mffssd}
J.~Zhang, L.~Zhu, L.~Xu, and Q.~Xie, ``Mffssd: An enhanced ssd for underwater
  object detection,'' in \emph{2020 Chinese Automation Congress (CAC)}.\hskip
  1em plus 0.5em minus 0.4em\relax IEEE, 2020, pp. 5938--5943.

\bibitem{lin2020roimix}
W.-H. Lin, J.-X. Zhong, S.~Liu, T.~Li, and G.~Li, ``Roimix: Proposal-fusion
  among multiple images for underwater object detection,'' in \emph{ICASSP
  2020-2020 IEEE International Conference on Acoustics, Speech and Signal
  Processing (ICASSP)}.\hskip 1em plus 0.5em minus 0.4em\relax IEEE, 2020, pp.
  2588--2592.

\bibitem{karimanzira2020object}
D.~Karimanzira, H.~Renkewitz, D.~Shea, and J.~Albiez, ``Object detection in
  sonar images,'' \emph{Electronics}, vol.~9, no.~7, p. 1180, 2020.

\bibitem{sung2020realistic}
M.~Sung, J.~Kim, M.~Lee, B.~Kim, T.~Kim, J.~Kim, and S.-C. Yu, ``Realistic
  sonar image simulation using deep learning for underwater object detection,''
  \emph{International Journal of Control, Automation and Systems}, vol.~18,
  no.~3, pp. 523--534, 2020.

\bibitem{yang2021research}
H.~Yang, P.~Liu, Y.~Hu, and J.~Fu, ``Research on underwater object recognition
  based on yolov3,'' \emph{Microsystem Technologies}, vol.~27, pp. 1837--1844,
  2021.

\bibitem{pan2021multi}
T.-S. Pan, H.-C. Huang, J.-C. Lee, and C.-H. Chen, ``Multi-scale resnet for
  real-time underwater object detection,'' \emph{Signal, Image and Video
  Processing}, vol.~15, pp. 941--949, 2021.

\bibitem{jiang2021underwater}
L.~Jiang, Y.~Wang, Q.~Jia, S.~Xu, Y.~Liu, X.~Fan, H.~Li, R.~Liu, X.~Xue, and
  R.~Wang, ``Underwater species detection using channel sharpening attention,''
  in \emph{Proceedings of the 29th ACM International Conference on Multimedia},
  2021, pp. 4259--4267.

\bibitem{yeh2021lightweight}
C.-H. Yeh, C.-H. Lin, L.-W. Kang, C.-H. Huang, M.-H. Lin, C.-Y. Chang, and
  C.-C. Wang, ``Lightweight deep neural network for joint learning of
  underwater object detection and color conversion,'' \emph{IEEE Transactions
  on Neural Networks and Learning Systems}, vol.~33, no.~11, pp. 6129--6143,
  2021.

\bibitem{chen2022swipenet}
L.~Chen, F.~Zhou, S.~Wang, J.~Dong, N.~Li, H.~Ma, X.~Wang, and H.~Zhou,
  ``Swipenet: Object detection in noisy underwater scenes,'' \emph{Pattern
  Recognition}, vol. 132, p. 108926, 2022.

\bibitem{alla2022vision}
D.~N.~V. Alla, V.~B.~N. Jyothi, H.~Venkataraman, and G.~Ramadass,
  ``Vision-based deep learning algorithm for underwater object detection and
  tracking,'' in \emph{OCEANS 2022-Chennai}.\hskip 1em plus 0.5em minus
  0.4em\relax IEEE, 2022, pp. 1--6.

\bibitem{zhang2022object}
L.~Zhang, C.~Li, and H.~Sun, ``Object detection/tracking toward underwater
  photographs by remotely operated vehicles (rovs),'' \emph{Future Generation
  Computer Systems}, vol. 126, pp. 163--168, 2022.

\bibitem{cai2022underwater}
S.~Cai, G.~Li, and Y.~Shan, ``Underwater object detection using collaborative
  weakly supervision,'' \emph{Computers and Electrical Engineering}, vol. 102,
  p. 108159, 2022.

\bibitem{jia2022underwater}
J.~Jia, M.~Fu, X.~Liu, and B.~Zheng, ``Underwater object detection based on
  improved efficientdet,'' \emph{Remote Sensing}, vol.~14, no.~18, p. 4487,
  2022.

\bibitem{wang2022underwater}
Y.~Wang, J.~Guo, and W.~He, ``Underwater object detection aided by image
  reconstruction,'' in \emph{2022 IEEE 24th International Workshop on
  Multimedia Signal Processing (MMSP)}.\hskip 1em plus 0.5em minus 0.4em\relax
  IEEE, 2022, pp. 1--6.

\bibitem{liang2022excavating}
X.~Liang and P.~Song, ``Excavating roi attention for underwater object
  detection,'' in \emph{2022 IEEE International Conference on Image Processing
  (ICIP)}.\hskip 1em plus 0.5em minus 0.4em\relax IEEE, 2022, pp. 2651--2655.

\bibitem{chen2023htdet}
G.~Chen, Z.~Mao, K.~Wang, and J.~Shen, ``Htdet: A hybrid transformer-based
  approach for underwater small object detection,'' \emph{Remote Sensing},
  vol.~15, no.~4, p. 1076, 2023.

\bibitem{song2023boosting}
P.~Song, P.~Li, L.~Dai, T.~Wang, and Z.~Chen, ``Boosting r-cnn: Reweighting
  r-cnn samples by rpn’s error for underwater object detection,''
  \emph{Neurocomputing}, 2023.

\bibitem{dai2023edge}
L.~Dai, H.~Liu, P.~Song, H.~Tang, R.~Ding, and S.~Li, ``Edge-guided
  representation learning for underwater object detection,'' \emph{CAAI
  Transactions on Intelligence Technology}, 2023.

\bibitem{fu2023learning}
C.~Fu, X.~Fan, J.~Xiao, W.~Yuan, R.~Liu, and Z.~Luo, ``Learning
  heavily-degraded prior for underwater object detection,'' \emph{IEEE
  Transactions on Circuits and Systems for Video Technology}, 2023.

\bibitem{zhou2024amsp}
J.~Zhou, Z.~He, K.-M. Lam, Y.~Wang, W.~Zhang, C.~Guo, and C.~Li, ``Amsp-uod:
  When vortex convolution and stochastic perturbation meet underwater object
  detection,'' in \emph{Proceedings of the AAAI Conference on Artificial
  Intelligence}, vol.~38, no.~7, 2024, pp. 7659--7667.

\bibitem{guo2024lightweight}
A.~Guo, K.~Sun, and Z.~Zhang, ``A lightweight yolov8 integrating fasternet for
  real-time underwater object detection,'' \emph{Journal of Real-Time Image
  Processing}, vol.~21, no.~2, pp. 1--15, 2024.

\bibitem{gao2024pe}
J.~Gao, Y.~Zhang, X.~Geng, H.~Tang, and U.~A. Bhatti, ``Pe-transformer: Path
  enhanced transformer for improving underwater object detection,''
  \emph{Expert Systems with Applications}, vol. 246, p. 123253, 2024.

\bibitem{ge2024advanced}
L.~Ge, P.~Singh, and A.~Sadhu, ``Advanced deep learning framework for
  underwater object detection with multibeam forward-looking sonar,''
  \emph{Structural Health Monitoring}, p. 14759217241235637, 2024.

\bibitem{dai2024gated}
L.~Dai, H.~Liu, P.~Song, and M.~Liu, ``A gated cross-domain collaborative
  network for underwater object detection,'' \emph{Pattern Recognition}, vol.
  149, p. 110222, 2024.

\bibitem{wang2024dual}
B.~Wang, Z.~Wang, W.~Guo, and Y.~Wang, ``A dual-branch joint learning network
  for underwater object detection,'' \emph{Knowledge-Based Systems}, vol. 293,
  p. 111672, 2024.

\bibitem{girshick2015fast}
R.~Girshick, ``Fast r-cnn,'' in \emph{Proceedings of the IEEE International
  Conference on Computer Vision}, 2015, pp. 1440--1448.

\bibitem{gavsparovic2022deep}
B.~Ga{\v{s}}parovi{\'c}, J.~Lerga, G.~Mau{\v{s}}a, and M.~Iva{\v{s}}i{\'c}-Kos,
  ``Deep learning approach for objects detection in underwater pipeline
  images,'' \emph{Applied Artificial Intelligence}, vol.~36, no.~1, p. 2146853,
  2022.

\bibitem{farhadi2018yolov3}
A.~Farhadi and J.~Redmon, ``Yolov3: An incremental improvement,'' in
  \emph{Computer Vision and Pattern Recognition}, vol. 1804.\hskip 1em plus
  0.5em minus 0.4em\relax Springer Berlin/Heidelberg, Germany, 2018, pp. 1--6.

\bibitem{boom2012supporting}
B.~J. Boom, P.~X. Huang, J.~He, and R.~B. Fisher, ``Supporting ground-truth
  annotation of image datasets using clustering,'' in \emph{Proceedings of the
  21st International Conference on Pattern Recognition (ICPR2012)}.\hskip 1em
  plus 0.5em minus 0.4em\relax IEEE, 2012, pp. 1542--1545.

\bibitem{beyan2013detecting}
C.~Beyan and R.~B. Fisher, ``Detecting abnormal fish trajectories using
  clustered and labeled data,'' in \emph{2013 IEEE International Conference on
  Image Processing}.\hskip 1em plus 0.5em minus 0.4em\relax IEEE, 2013, pp.
  1476--1480.

\bibitem{kavasidis2014innovative}
I.~Kavasidis, S.~Palazzo, R.~D. Salvo, D.~Giordano, and C.~Spampinato, ``An
  innovative web-based collaborative platform for video annotation,''
  \emph{Multimedia Tools and Applications}, vol.~70, pp. 413--432, 2014.

\bibitem{liu2021new}
C.~Liu, Z.~Wang, S.~Wang, T.~Tang, Y.~Tao, C.~Yang, H.~Li, X.~Liu, and X.~Fan,
  ``A new dataset, poisson gan and aquanet for underwater object grabbing,''
  \emph{IEEE Transactions on Circuits and Systems for Video Technology},
  vol.~32, no.~5, pp. 2831--2844, 2021.

\bibitem{fabbri2018enhancing}
C.~Fabbri, M.~J. Islam, and J.~Sattar, ``Enhancing underwater imagery using
  generative adversarial networks,'' in \emph{2018 IEEE International
  Conference on Robotics and Automation (ICRA)}.\hskip 1em plus 0.5em minus
  0.4em\relax IEEE, 2018, pp. 7159--7165.

\bibitem{chen2019towards}
X.~Chen, J.~Yu, S.~Kong, Z.~Wu, X.~Fang, and L.~Wen, ``Towards real-time
  advancement of underwater visual quality with gan,'' \emph{IEEE Transactions
  on Industrial Electronics}, vol.~66, no.~12, pp. 9350--9359, 2019.

\bibitem{liu2020real}
R.~Liu, X.~Fan, M.~Zhu, M.~Hou, and Z.~Luo, ``Real-world underwater
  enhancement: Challenges, benchmarks, and solutions under natural light,''
  \emph{IEEE Transactions on Circuits and Systems for Video Technology},
  vol.~30, no.~12, pp. 4861--4875, 2020.

\bibitem{chen2020perceptual}
L.~Chen, Z.~Jiang, L.~Tong, Z.~Liu, A.~Zhao, Q.~Zhang, J.~Dong, and H.~Zhou,
  ``Perceptual underwater image enhancement with deep learning and physical
  priors,'' \emph{IEEE Transactions on Circuits and Systems for Video
  Technology}, 2020.

\bibitem{liu2022twin}
R.~Liu, Z.~Jiang, S.~Yang, and X.~Fan, ``Twin adversarial contrastive learning
  for underwater image enhancement and beyond,'' \emph{IEEE Transactions on
  Image Processing}, vol.~31, pp. 4922--4936, 2022.

\bibitem{dai2021attentional}
Y.~Dai, F.~Gieseke, S.~Oehmcke, Y.~Wu, and K.~Barnard, ``Attentional feature
  fusion,'' in \emph{Proceedings of the IEEE/CVF Winter Conference on
  Applications of Computer Vision}, 2021, pp. 3560--3569.

\bibitem{zhang2021lightweight}
M.~Zhang, S.~Xu, W.~Song, Q.~He, and Q.~Wei, ``Lightweight underwater object
  detection based on yolo v4 and multi-scale attentional feature fusion,''
  \emph{Remote Sensing}, vol.~13, no.~22, p. 4706, 2021.

\bibitem{singh2018analysis}
B.~Singh and L.~S. Davis, ``An analysis of scale invariance in object detection
  snip,'' in \emph{Proceedings of the IEEE Conference on Computer Vision and
  Pattern Recognition}, 2018, pp. 3578--3587.

\bibitem{najibi2019autofocus}
M.~Najibi, B.~Singh, and L.~S. Davis, ``Autofocus: Efficient multi-scale
  inference,'' in \emph{Proceedings of the IEEE/CVF International Conference on
  Computer Vision}, 2019, pp. 9745--9755.

\bibitem{yang2019scrdet}
X.~Yang, J.~Yang, J.~Yan, Y.~Zhang, T.~Zhang, Z.~Guo, X.~Sun, and K.~Fu,
  ``Scrdet: Towards more robust detection for small, cluttered and rotated
  objects,'' in \emph{Proceedings of the IEEE/CVF International Conference on
  Computer Vision}, 2019, pp. 8232--8241.

\bibitem{gao2023global}
T.~Gao, Q.~Niu, J.~Zhang, T.~Chen, S.~Mei, and A.~Jubair, ``Global to local: A
  scale-aware network for remote sensing object detection,'' \emph{IEEE
  Transactions on Geoscience and Remote Sensing}, 2023.

\bibitem{zhang2023superyolo}
J.~Zhang, J.~Lei, W.~Xie, Z.~Fang, Y.~Li, and Q.~Du, ``Superyolo: Super
  resolution assisted object detection in multimodal remote sensing imagery,''
  \emph{IEEE Transactions on Geoscience and Remote Sensing}, vol.~61, pp.
  1--15, 2023.

\bibitem{rabbi2020small}
J.~Rabbi, N.~Ray, M.~Schubert, S.~Chowdhury, and D.~Chao, ``Small-object
  detection in remote sensing images with end-to-end edge-enhanced gan and
  object detector network,'' \emph{Remote Sensing}, vol.~12, no.~9, p. 1432,
  2020.

\bibitem{bashir2021small}
S.~M.~A. Bashir and Y.~Wang, ``Small object detection in remote sensing images
  with residual feature aggregation-based super-resolution and object detector
  network,'' \emph{Remote Sensing}, vol.~13, no.~9, p. 1854, 2021.

\bibitem{song2022learning}
H.~Song, M.~Kim, D.~Park, Y.~Shin, and J.-G. Lee, ``Learning from noisy labels
  with deep neural networks: A survey,'' \emph{IEEE Transactions on Neural
  Networks and Learning Systems}, 2022.

\bibitem{song2019selfie}
H.~Song, M.~Kim, and J.-G. Lee, ``Selfie: Refurbishing unclean samples for
  robust deep learning,'' in \emph{International Conference on Machine
  Learning}.\hskip 1em plus 0.5em minus 0.4em\relax PMLR, 2019, pp. 5907--5915.

\bibitem{wang2018iterative}
Y.~Wang, W.~Liu, X.~Ma, J.~Bailey, H.~Zha, L.~Song, and S.-T. Xia, ``Iterative
  learning with open-set noisy labels,'' in \emph{Proceedings of the IEEE
  Conference on Computer Vision and Pattern Recognition}, 2018, pp. 8688--8696.

\bibitem{patel2023adaptive}
D.~Patel and P.~Sastry, ``Adaptive sample selection for robust learning under
  label noise,'' in \emph{Proceedings of the IEEE/CVF Winter Conference on
  Applications of Computer Vision}, 2023, pp. 3932--3942.

\bibitem{wang2021proselflc}
X.~Wang, Y.~Hua, E.~Kodirov, D.~A. Clifton, and N.~M. Robertson, ``Proselflc:
  Progressive self label correction for training robust deep neural networks,''
  in \emph{Proceedings of the IEEE/CVF Conference on Computer Vision and
  Pattern Recognition}, 2021, pp. 752--761.

\bibitem{ma2023ctw}
P.~Ma, Z.~Liu, J.~Zheng, L.~Wang, and Q.~Ma, ``Ctw: confident time-warping for
  time-series label-noise learning,'' in \emph{Proceedings of the Thirty-Second
  International Joint Conference on Artificial Intelligence}, 2023, pp.
  4046--4054.

\bibitem{wei2020combating}
H.~Wei, L.~Feng, X.~Chen, and B.~An, ``Combating noisy labels by agreement: A
  joint training method with co-regularization,'' in \emph{Proceedings of the
  IEEE/CVF Conference on Computer Vision and Pattern Recognition}, 2020, pp.
  13\,726--13\,735.

\bibitem{yao2020searching}
Q.~Yao, H.~Yang, B.~Han, G.~Niu, and J.~T.-Y. Kwok, ``Searching to exploit
  memorization effect in learning with noisy labels,'' in \emph{International
  Conference on Machine Learning}.\hskip 1em plus 0.5em minus 0.4em\relax PMLR,
  2020, pp. 10\,789--10\,798.

\bibitem{li2017learning}
Y.~Li, J.~Yang, Y.~Song, L.~Cao, J.~Luo, and L.-J. Li, ``Learning from noisy
  labels with distillation,'' in \emph{Proceedings of the IEEE International
  Conference on Computer Vision}, 2017, pp. 1910--1918.

\bibitem{wu2021ngc}
Z.-F. Wu, T.~Wei, J.~Jiang, C.~Mao, M.~Tang, and Y.-F. Li, ``Ngc: A unified
  framework for learning with open-world noisy data,'' in \emph{Proceedings of
  the IEEE/CVF International Conference on Computer Vision}, 2021, pp. 62--71.

\bibitem{ghosh2017robust}
A.~Ghosh, H.~Kumar, and P.~S. Sastry, ``Robust loss functions under label noise
  for deep neural networks,'' in \emph{Proceedings of the AAAI Conference on
  Artificial Intelligence}, vol.~31, no.~1, 2017.

\bibitem{zhang2018generalized}
Z.~Zhang and M.~Sabuncu, ``Generalized cross entropy loss for training deep
  neural networks with noisy labels,'' \emph{Advances in Neural Information
  Processing Systems}, vol.~31, 2018.

\bibitem{zhou2023asymmetric}
X.~Zhou, X.~Liu, D.~Zhai, J.~Jiang, and X.~Ji, ``Asymmetric loss functions for
  noise-tolerant learning: Theory and applications,'' \emph{IEEE Transactions
  on Pattern Analysis and Machine Intelligence}, 2023.

\bibitem{chang2021image}
N.~Chang, Z.~Yu, Y.-X. Wang, A.~Anandkumar, S.~Fidler, and J.~M. Alvarez,
  ``Image-level or object-level? a tale of two resampling strategies for
  long-tailed detection,'' in \emph{International Conference on Machine
  Learning}.\hskip 1em plus 0.5em minus 0.4em\relax PMLR, 2021, pp. 1463--1472.

\bibitem{feng2021exploring}
C.~Feng, Y.~Zhong, and W.~Huang, ``Exploring classification equilibrium in
  long-tailed object detection,'' in \emph{Proceedings of the IEEE/CVF
  International Conference on Computer Vision}, 2021, pp. 3417--3426.

\bibitem{ren2018learning}
M.~Ren, W.~Zeng, B.~Yang, and R.~Urtasun, ``Learning to reweight examples for
  robust deep learning,'' in \emph{International Conference on Machine
  Learning}.\hskip 1em plus 0.5em minus 0.4em\relax PMLR, 2018, pp. 4334--4343.

\bibitem{mahajan2018exploring}
D.~Mahajan, R.~Girshick, V.~Ramanathan, K.~He, M.~Paluri, Y.~Li, A.~Bharambe,
  and L.~Van Der~Maaten, ``Exploring the limits of weakly supervised
  pretraining,'' in \emph{Proceedings of the European Conference on Computer
  Vision (ECCV)}, 2018, pp. 181--196.

\bibitem{cui2019class}
Y.~Cui, M.~Jia, T.-Y. Lin, Y.~Song, and S.~Belongie, ``Class-balanced loss
  based on effective number of samples,'' in \emph{Proceedings of the IEEE
  Conference on Computer Vision and Pattern Recognition}, 2019, pp. 9268--9277.

\bibitem{li2020overcoming}
Y.~Li, T.~Wang, B.~Kang, S.~Tang, C.~Wang, J.~Li, and J.~Feng, ``Overcoming
  classifier imbalance for long-tail object detection with balanced group
  softmax,'' in \emph{Proceedings of the IEEE/CVF Conference on Computer Vision
  and Pattern Recognition}, 2020, pp. 10\,991--11\,000.

\bibitem{qi2023balanced}
T.~Qi, H.~Xie, P.~Li, J.~Ge, and Y.~Zhang, ``Balanced classification: A unified
  framework for long-tailed object detection,'' \emph{IEEE Transactions on
  Multimedia}, 2023.

\bibitem{fisher2016fish4knowledge}
R.~B. Fisher, Y.-H. Chen-Burger, D.~Giordano, L.~Hardman, F.-P. Lin
  \emph{et~al.}, \emph{Fish4Knowledge: collecting and analyzing massive coral
  reef fish video data}.\hskip 1em plus 0.5em minus 0.4em\relax Springer, 2016,
  vol. 104.

\bibitem{wang2019region}
J.~Wang, K.~Chen, S.~Yang, C.~C. Loy, and D.~Lin, ``Region proposal by guided
  anchoring,'' in \emph{Proceedings of the IEEE/CVF Conference on Computer
  Vision and Pattern Recognition}, 2019, pp. 2965--2974.

\bibitem{cai2019cascade}
Z.~Cai and N.~Vasconcelos, ``Cascade r-cnn: High quality object detection and
  instance segmentation,'' \emph{IEEE Transactions on Pattern Analysis and
  Machine Intelligence}, vol.~43, no.~5, pp. 1483--1498, 2019.

\bibitem{kong2020foveabox}
T.~Kong, F.~Sun, H.~Liu, Y.~Jiang, L.~Li, and J.~Shi, ``Foveabox: Beyound
  anchor-based object detection,'' \emph{IEEE Transactions on Image
  Processing}, vol.~29, pp. 7389--7398, 2020.

\bibitem{zhang2020bridging}
S.~Zhang, C.~Chi, Y.~Yao, Z.~Lei, and S.~Z. Li, ``Bridging the gap between
  anchor-based and anchor-free detection via adaptive training sample
  selection,'' in \emph{Proceedings of the IEEE/CVF Conference on Computer
  Vision and Pattern Recognition}, 2020, pp. 9759--9768.

\bibitem{qiao2021detectors}
S.~Qiao, L.-C. Chen, and A.~Yuille, ``Detectors: Detecting objects with
  recursive feature pyramid and switchable atrous convolution,'' in
  \emph{Proceedings of the IEEE/CVF Conference on Computer Vision and Pattern
  Recognition}, 2021, pp. 10\,213--10\,224.

\bibitem{lu2019grid}
X.~Lu, B.~Li, Y.~Yue, Q.~Li, and J.~Yan, ``Grid r-cnn,'' in \emph{Proceedings
  of the IEEE/CVF Conference on Computer Vision and Pattern Recognition}, 2019,
  pp. 7363--7372.

\bibitem{lin2014microsoft}
T.-Y. Lin, M.~Maire, S.~Belongie, J.~Hays, P.~Perona, D.~Ramanan,
  P.~Doll{\'a}r, and C.~L. Zitnick, ``Microsoft coco: Common objects in
  context,'' in \emph{Computer Vision--ECCV 2014: 13th European Conference,
  Zurich, Switzerland, September 6-12, 2014, Proceedings, Part V 13}.\hskip 1em
  plus 0.5em minus 0.4em\relax Springer, 2014, pp. 740--755.

\end{thebibliography}

\ifCLASSOPTIONcaptionsoff
\vspace{-15 mm}

\begin{IEEEbiographynophoto}{Long Chen} received the B.S. degree in Northeast Normal University in 2013 and the M.S. degree in Computer Architecture at the Vision Lab of Ocean University of China in 2018. He achieved the PhD degree in School of Computing and Mathematical Sciences, University of Leicester in 2022. Currently, he is a postdoc with the Department of Medical Physics and Biomedical Engineering, University College London (UCL). His general research interests are in the area of deep learning, machine learning, and computer vision.
\end{IEEEbiographynophoto}

\fi

\end{document}